\documentclass[journal]{IEEEtran}

\usepackage{moreverb,url}
\usepackage{color, colortbl}
\usepackage[labelfont=bf,textfont=it,hypcap=true,labelsep=space]{caption,subcaption}
\usepackage{booktabs}
\usepackage{array,multirow,graphicx} 
\usepackage{bbding} 
\usepackage{amsmath} 
\usepackage{amssymb}  
\usepackage{floatrow}
\usepackage{algorithm,algpseudocode}
\usepackage{caption}
\usepackage{subcaption}
\usepackage[normalem]{ulem}
\usepackage[hang]{footmisc}
\usepackage[colorlinks,bookmarksopen,bookmarksnumbered,citecolor=red,urlcolor=red, unicode]{hyperref}
\usepackage{cleveref}
\usepackage{xcolor}
\usepackage[utf8]{inputenc}

\usepackage{footnote}
\usepackage{flushend}

\definecolor{d_gray}{gray}{0.05}
\definecolor{l_gray}{gray}{0.95}
\definecolor{ll_gray}{gray}{0.97}

\newcommand{\etal}{\textit{et al.}}

\begin{document}

\title{MODS -- A USV-oriented object detection and obstacle segmentation benchmark}

\author{Borja~Bovcon$^*$, Jon~Muhovič$^*$, Duško~Vranac, Dean~Mozetič, Janez~Perš, Matej~Kristan~\IEEEmembership{Member,~IEEE,}
\thanks{B. Bovcon and M. Kristan are with University of Ljubljana, Faculty of Computer and Information Science.}
\thanks{J. Muhovič and J. Perš are with University of Ljubljana, Faculty of Electrical Engineering.}
\thanks{D. Vranac is with MPRI and D. Mozetič is with Sirio d.o.o.}
\thanks{$*$ These authors contributed equally.}}

\markboth{IEEE Transactions on Intelligent Transportation Systems}%
{Bovcon, Muhovič \MakeLowercase{\textit{et al.}}: MODS -- A USV-oriented object detection and obstacle segmentation benchmark}

\maketitle

\begin{abstract}
Small-sized unmanned surface vehicles (USV) are coastal water devices with a broad range of applications such as environmental control and surveillance.  A crucial capability for autonomous operation is obstacle detection for timely reaction and collision avoidance, which has been recently explored in the context of camera-based visual scene interpretation. Owing to curated datasets, substantial advances in scene interpretation have been made in a related field of unmanned ground vehicles. However, the current maritime datasets do not adequately capture the complexity of real-world USV scenes and the evaluation protocols are not standardised, which makes cross-paper comparison of different methods difficult and hinders the progress. To address these issues, we introduce a new obstacle detection benchmark MODS, which considers two major perception tasks: maritime object detection and the more general maritime obstacle segmentation. We present a new diverse maritime evaluation dataset containing approximately 81k stereo images synchronized with an on-board IMU, with over 60k objects annotated. We propose a new obstacle segmentation performance evaluation protocol that reflects the detection accuracy in a way meaningful for practical USV navigation. Nineteen recent state-of-the-art  object detection and obstacle segmentation methods are evaluated using the proposed protocol, creating a benchmark to facilitate development of the field. The proposed dataset, as well as evaluation routines, are made publicly available at \url{vicos.si/resources}.
\end{abstract}


%
\IEEEpeerreviewmaketitle

\section{Introduction}

\IEEEPARstart{A}{utonomy} is transforming most industries, and maritime robotics is no exception. In the last decade, the interest in autonomous unmanned surface vehicles (USV) has escalated. With 90\% of goods moved across the world in vessels, autonomous boats and ships present a considerable market opportunity.
Autonomous ships and ferries are expected to significantly reduce the environmental hazard and emissions and small-sized ($\sim$2m) USVs are expected to contribute to green coastal communities. The small-sized USVs, in particular, present affordable devices for automated inspection of hazardous areas and periodic surveillance of coastal waters and have a strong potential for a wide-spread use.

\begin{figure}
    \centering
        \includegraphics[width=\columnwidth]{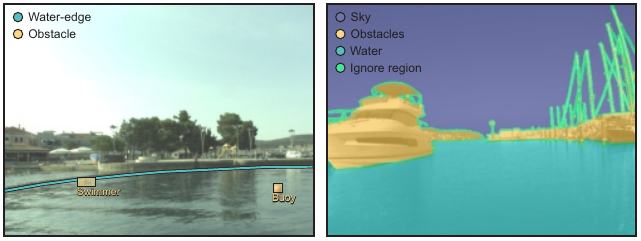}
    \caption{Two major perception tasks crucial for USV collision avoidance: maritime object detection (left) and obstacle segmentation (right). The first problem assumes that static obstacles such as shoreline are known in advance by maps, while the second problem addresses prediction of the entire navigable area in which obstacle localization is implicit.}
\label{fig:anchors}
\end{figure}

Autonomous vehicles are complex machines, usually comprised of several interconnected modules: a sensing module with array of sensors (one or more cameras and possibly RADAR~\cite{Onunka2010}, LIDAR~\cite{Jimenez2009} or SONAR~\cite{Heidarsson2011}), a perception module which evaluates and fuses sensor information~\cite{Muresan2019}, the path planning module~\cite{Kim2017} which relies on the output of the perception module, and finally, the control module, which controls the actuators according to the planned path~\cite{Arikan2018}. 

Embodied in the perception module, a kingpin of the USV autonomy is the environment perception capability, which still lags far behind the control and hardware research~\cite{prasad2018overview,jorge2019survey}. Safe navigation in dynamic environments such as coastal waters requires robust obstacle detection for timely reaction and collision avoidance. Perception, however, is limited by the physical constraints of small-sized USVs, which prohibits the use of heavy, power consuming sensors such as RADAR and LIDAR. Cameras as light-weight, low-power and information rich sensors have thus attracted considerable attention of the research community~\cite{zhang2020review}.
 
From the growing body of research in camera-based collision avoidance for USVs, two core perception problems emerged. The first problem, called \textit{maritime object detection}, is used in setups in which one can reasonably assume that static obstacles such as piers and shore are known in advance and can be avoided by using GPS and shoreline maps. Thus only dynamic obstacles from known classes such as ships, boats, swimmers and buoys have to be detected to ensure safe navigation. This problem has been traditionally addressed by saliency-based region detectors~\cite{albrecht2011visual_saliency1,liu2008omni,makantasis2013vision_saliency3,sobral2015double_saliency4,cane2016saliency_saliency5,akilan2019sendec}, stereo detectors~\cite{wang2013stereovision,muhovic2019obstacle} and most recently the research has focused on specialized convolutional neural network (CNN) object detectors~\cite{lee2018image,moosbauer2019benchmark_smd_instance,yang2019surface}.

The second problem addresses situations in which shoreline maps do not exist, or are not sufficiently accurate for reliable static obstacle detection. The problem is called \textit{obstacle segmentation}, since the task here is to segment the entire input image into navigable and non-navigable surface, i.e., assign to each pixel a label indicating whether it belongs to an obstacle, water or sky. This problem is currently the dominant paradigm in camera-based unmanned ground vehicles (UGV) semantic scene understanding~\cite{paszke2016enet,badrinarayanan2017segnet,zhao2017pyramid,lin2017refinenet,yu2018bisenet,chen2018deeplab3,xu2020salmnet} and is gaining traction in the USV research as well~\cite{KristanCYB2015,cane2018evaluating,kim2019vision,steccanella2020intcatch,bovcon2020wasr_icra}. In addition, multimodal deep-learning-based UGV methods have recently emerged that utilise information from various sensors (e.g., thermal cameras~\cite{sun2020fuseseg}, depth-cameras~\cite{wang2019self}, near-infrared cameras~\cite{luo2010pedestrian}) to improve the predicted segmentation accuracy. This trend, however, is still relatively new in the domain of USVs and only a couple of methods exploit it~\cite{wang2013stereovision,muhovic2019obstacle,bovcon2018stereo,bovcon2020wasr_icra}.

Maritime object detection and obstacle segmentation for USVs have been the focus of many recent works. But the proposed methods are trained on different, often private, datasets using non-standardised performance evaluation measures. This makes objective cross-paper comparison of the different methods difficult and slows down the pace of advancements. In related fields such as UGV visual scene interpretation~\cite{geiger2012we_kitti,cordts2016cityscapes,Maddern2017}, multi-sensor UGV localization~\cite{Barnes2020} and visual object tracking~\cite{VOT2019,Dendorfer2020}, the progress has been driven by carefully curated benchmarks and challenges on which best practices and perception algorithm designs can be evaluated under a common setup. Attempts have been recently made in this direction in USV literature~\cite{prasad2018overview,cane2018evaluating,moosbauer2019benchmark_smd_instance,bovcon2019mastr}, but a comprehensive USV-specific benchmark with sufficiently large and diverse evaluation dataset and protocols is still missing. 

To address the above issues, we propose a large unified benchmark -- MODS -- designed for evaluation of classical object detection methods as well as segmentation-based obstacle detection methods. We make the following three contributions: (i) a new carefully annotated IMU-synchronized stereo RGB dataset captured under various weather conditions in which the USV is expected to operate, (ii) an obstacle segmentation performance evaluation protocol and (iii) performance evaluation of popular obstacle detection approaches. The presented dataset is the largest of its kind, amounting to over $80$k frames with over $60$k dynamic obstacles annotated by bounding boxes and static obstacles such as shores and piers annotated by water-obstacle boundaries. Curated training datasets for learning-based object detection and obstacle segmentation methods are specified as well. The proposed obstacle segmentation performance evaluation protocol is implemented in Python and it measures the detection accuracy meaningful for practical USV navigation. A comprehensive set of popular state-of-the-art approaches for USV maritime object detection and obstacle segmentation are benchmarked using the new protocol and the dataset.
The datasets and routines implementing the evaluation protocols are publicly available to push towards standardised performance evaluation in camera-based USV obstacle detection methods, facilitate cross-paper results comparison and accelerate future research in this growing field.
 
The remainder of the paper is structured as follows. Section~\ref{sec:related_work} overviews the state of USV obstacle detection and related works in benchmarking. Section~\ref{sec:acquisition_hardware} describes the dataset acquisition and annotation, and the proposed performance evaluation protocols are described in Section~\ref{sec:performance_measures}. The new benchmark is reported in Section~\ref{sec:benchmark} and conclusions are drawn in Section~\ref{sec:conclusion}.

\section{Related work}
\label{sec:related_work}
In the following we first briefly overview the main historical development of maritime object detection and obstacle segmentation methods (Section~\ref{subsec:related_work_detection_methods}) and then focus on related maritime datasets proposed for evaluation and training of these algorithms (Section~\ref{subsec:related_work_datasets}).

\subsection{Maritime detection and segmentation}
\label{subsec:related_work_detection_methods}
Several works,~\cite{liu2008omni,albrecht2011visual_saliency1,makantasis2013vision_saliency3,sobral2015double_saliency4,cane2016saliency_saliency5}, consider maritime object detection as a saliency estimation problem. These methods assume the objects are well-distinguishable from their immediate background. This assumption is violated in many situations like in presence of fog and glitter and for objects that are visually similar to water, leading to failure. Classical background subtraction methods are also not suitable for USVs since the undulating sea continuously rocks the USV and violates the static camera assumption, which causes a high false positive rate~\cite{prasad2018overview,chan2020comprehensive}. More advanced foreground extraction methods based on deep learning (e.g.~\cite{akilan2019sendec}) address this drawback, however, they still tend to fail in the presence of visual ambiguities.

Wang~\etal~\cite{wang2013stereovision} proposed a stereo obstacle detection method to reduce the visual ambiguity by fitting a 3D plane to the estimated depth map and detecting obstacles as entities protruding the plane. The method detects the obstacles at close range, but fails at long range or on obstacles that do not significantly protrude the water surface (i.e., flat floating debris and swimmers). Another drawback is that stereo systems fail in calm waters. Muhovič~\etal~\cite{muhovic2019obstacle} addressed this drawback by using an on-board IMU to correctly approximate the water surface even in the state of a calm sea.

Deep convolutional neural networks (CNNs) enable extraction of rich visual features and are thus gaining popularity in maritime perception methods. Several works \cite{lee2018image,moosbauer2019benchmark_smd_instance,yang2019surface} applied general-purpose detection CNNs~\cite{ren2015faster,he2017mask} to the aquatic domain to detect and classify different types of ships. \cite{ma2019convolutional} extended Faster R-CNN classifier with ResNet backbone and added trainable weight parameters to skip connections in DenseNet block~\cite{huang2017densely}. Although these methods achieve state-of-the-art classification results, they are unable to detect static obstacles like piers and even general dynamic objects (such as debris) which were not included in the training phase.

Detection of general static and dynamic obstacles is principally addressed by segmentation-based methods. Kristan~\etal~\cite{KristanCYB2015} proposed a hand-crafted semantic segmentation method based on fitting a statistical structured model to the image. Their method was recently extended by \cite{bovcon2018stereo,bovcon2018iros} with IMU and a stereo-camera system fusion. Information from additional sensors contributed to a better water-edge estimation as well as reduced number of false positive detections caused by sun-glitter. These approaches rely on simple visual features incapable to robustly cope with a diverse aquatic environment, thus leading to poor segmentation, particularly in presence of mirroring and glitter.

Segmentation-based CNNs, on the other hand, substantially increase the detection robustness. Recent works from \cite{bovcon2018iros} and \cite{cane2018evaluating} report that general-purpose CNN segmentation architectures outperform maritime hand-crafted segmentation methods, but still perform poorly on mirroring and segmentation of small obstacles. Kim~\etal~\cite{kim2019vision} applied skip-connections and whitening layer to E-Net~\cite{paszke2016enet} to improve small obstacle detection. Steccanella~\etal~\cite{steccanella2020intcatch} proposed replacing traditional convolutional layers in U-Net~\cite{ronneberger2015u} with depth-wise convolutions to improve water segmentation. Alternatively, \cite{bovcon2020wasr_icra} proposed a non-symmetric encoder-decoder network with a ResNet101~\cite{he2016resnet} backbone that improves the overall segmentation accuracy by visual and inertial information fusion. Deep learning based methods, are, like in related domain of autonomous cars, becoming the dominant paradigm in maritime object and obstacle detection.

\subsection{Maritime visual perception datasets}
\label{subsec:related_work_datasets}

Several datasets have been proposed to evaluate various algorithms for maritime surveillance and robotic navigation. Fefilatyev~\etal~\cite{fefilatyev2006horizon_buoy_dataset} proposed a dataset with ten sequences recorded on the same day from a bouy looking at the open sea. The dataset is used for horizon detection evaluation and does not contain obstacles. The sequences were recorded in the same scene on the same day, which limits their visual diversity.  
Bloisi~\etal~\cite{mardct_dataset} proposed ten sequences for maritime object tracking. Visual diversity is increased by recording at different times of the day and several dynamic obstacles such as ships, boats and jet-skies are annotated. However, since all obstacles are dark on a very bright water, they pose little challenge for obstacle detection. Marques~\etal~\cite{seagull_dataset} and Ribeiro~\etal~\cite{ribeiro2017dataset_airborne_dataset} recorded two visually diverse datasets for maritime airborne-based detection. The dataset is designed for drone applications and it does not feature vantage point observed in autonomous boats. 

Patino~\etal~\cite{ipatch_dataset} presented a dataset with 14 multi-sensor sequences used for obstacle detection, tracking and threat recognition evaluation. The dataset contains annotations for horizon and dynamic obstacles, but does not contain small-sized obstacles like bouys.
Kristan~\etal~\cite{KristanCYB2015} introduced a maritime obstacle detection dataset with 12 diverse sequences captured from a USV, which was later extended by Bovcon~\etal~\cite{bovcon2018stereo} with 28 stereo camera sequences synchronized with an IMU. Both datasets were recorded in the same port and contain annotations for horizon, water-edge and large as well as small dynamic obstacles. Visual diversity was maintained by recording under different weather conditions.

Prasad~\etal~\cite{smd_prasad2017video} proposed a large maritime surveillance dataset containing 51 RGB and 30 NIR sequences recorded at different times of the day and under different weather conditions. Most of the sequences are recorded from a fixed on-shore point, while some are captured from a ship with  much higher vantage point than in robotic boats. Since it was designed primarily for surveillance, the scenes are very static with little motion. The dynamic obstacles and horizon are well annotated and recently~\cite{moosbauer2019benchmark_smd_instance} provided coarse instance segmentation labels calculated by a color-based semi-automatic method.
 
Gundogdu~\etal~\cite{marvel_dataset} proposed a dataset with 400k patches for vessel classification task, but the dataset cannot be used for detector evaluation since the vessel locations are not annotated. Soloviev~\etal~\cite{soloviev_detector_dataset} recently proposed a dataset of $\sim$2k images. The dataset is designed for evaluation of pre-trained vessel detectors, thus static obstacles like shore and dynamic obstacles such as bouys are not annotated. 

Most of the datasets have been proposed for evaluation of object detectors and only few were designed for training segmentation methods. Steccanella~\etal~\cite{steccanella2020intcatch} proposed per-pixel annotated dataset of 191 images recorded on seven separate occasions on a lake for training and testing deep segmentation methods. The dataset contains two semantic labels, i.e., water and non-water. The visual diversity is limited, the test set is not well separated from the training set and since the labels do not distinguish between sky and obstacles, only water segmentation quality can be evaluated. Currently the largest and most detailed dataset for training deep maritime segmentation networks was proposed by \cite{bovcon2019mastr}. The dataset was recorded during different times of the day and under various weather conditions over a span of two years and contains $\sim$1300 images with per-pixel labels for water, sky and obstacles.

The work reported in this paper complements the existing works~\cite{KristanCYB2015,ipatch_dataset,smd_prasad2017video,bovcon2018stereo,bovcon2019mastr,steccanella2020intcatch} by proposing a carefully annotated evaluation dataset, which is order of magnitude larger than the existing datasets (it consists of more than $80$k stereo images, while the second largest publicly available dataset, SMD~\cite{smd_prasad2017video}, has only $31$k images), with highly diverse obstacles (more than $60$k annotated objects) recorded under realistic surveillance conditions at various locations. 
Evaluation protocols and training datasets are developed as well to promote standardised and objective comparison of maritime object detectors as well as segmentation-based obstacle detectors. 
A wide range of state-of-the-art obstacle segmentation and object detection methods is evaluated, creating the most thoroughly constructed benchmark in the USV object and obstacle detection with a publicly available toolkit to date~\footnote{The evaluation routines and the datasets are available at \url{vicos.si/resources}.}.
Note that, compared to related benchmarks~\cite{bovcon2018stereo,bovcon2020wasr_icra,muhovic2019obstacle,steccanella2020intcatch,cane2018evaluating}, the most recent state-of-the-art obstacle detection methods achieve a significantly poorer performance on our benchmark, which also makes it currently the most challenging benchmark.

\section{Dataset acquisition hardware}
\label{sec:acquisition_hardware}
\begin{figure}
    \centering
        \includegraphics[width=0.95\columnwidth]{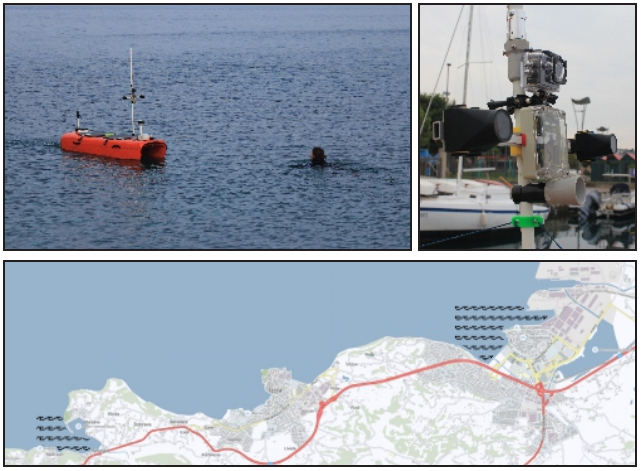}
    \caption{
    USV on a mission (upper left) and a close-up of its acquisition hardware (upper right). The bottom row shows a map of Slovenian coast with waves patterns denoting the dataset acquisition areas.
    }
\label{fig:acquisition_hardware}
\end{figure}

To reflect the realism of the USV environment encountered on an average mission, the dataset for our new benchmark was captured by a real small-sized USV, shown in Figure~\ref{fig:acquisition_hardware}. The USV uses a steerable thrust propeller for guidance, and can reach the maximum velocity of $2.5$m/s. It is equipped with LIDAR, compass, GPS, IMU unit, two side cameras and a main stereo camera system \emph{Vrmagic VRmMFC}, which comprise of two \emph{Vrmagic VRmS-14/C-COB} CCD sensors with baseline $0.3$m \emph{Thorlabs MVL4WA} lens with $3.5$mm focal length, maximum aperture of $\mathrm{f}/1.4$, and a $132.1^{\circ}$ FOV. 

The stereo system, shown in Figure~\ref{fig:acquisition_hardware} right, is mounted approximately $0.7$m above the water surface, faces forward and is z-axis aligned with the IMU. Cameras are connected to the on-board computer through USB-2.0 bus which restricts the data flow to $10$ frames per second at resolution $1,278 \times 958$ pixels. The aperture of cameras is automatically adjusted according to the lighting conditions, preventing underexposure and occurrence of indistinguishable dark areas in shades. The IMU is calibrated with the stereo system using IMU-camera calibration toolbox~\cite{bovcon2018stereo}.

\section{Dataset construction}\label{sec:dataset_construction}

Our primary goal was to construct a comprehensive dataset for evaluating the two classes of detection problems in USVs. On the other hand, many modern maritime object and obstacle segmentation methods rely on pre-trained deep learning backbones, which might require fine-tuning on maritime perception tasks. As a compromise, we thus dedicated most of our resources to acquisition and careful annotation of the evaluation dataset (Section~\ref{sec:evaluation_dataset}) and created training datasets by collecting and consolidating existing maritime datasets (Section~\ref{subsec:training_datasets}). This also ensured a reduced evaluation bias, since the CNNs would not overfit to the characteristics of the evaluation set.

\subsection{Evaluation dataset acquisition}\label{sec:evaluation_dataset}

A large corpus of initial sequences was acquired during eight voyages with our USV (Section~\ref{sec:acquisition_hardware}) over a span of seven months in the years 2018-2019. The sequences were captured in two geographically disjoint areas of Slovenian coastal waters (port of Koper and Strunjan -- see Figure~\ref{fig:acquisition_hardware}) to diversify the obstacles and environment appearance. To further diversify the dataset and capture the realism of USV missions, the voyages were planned at different times of the day and under different weather conditions. An expert manually piloted the USV and included realistic navigation scenarios with dangerous situations in which the boat is heading straight towards an obstacle or passing it by in a close range.

\begin{figure}
    \centering
        \includegraphics[width=0.95\columnwidth]{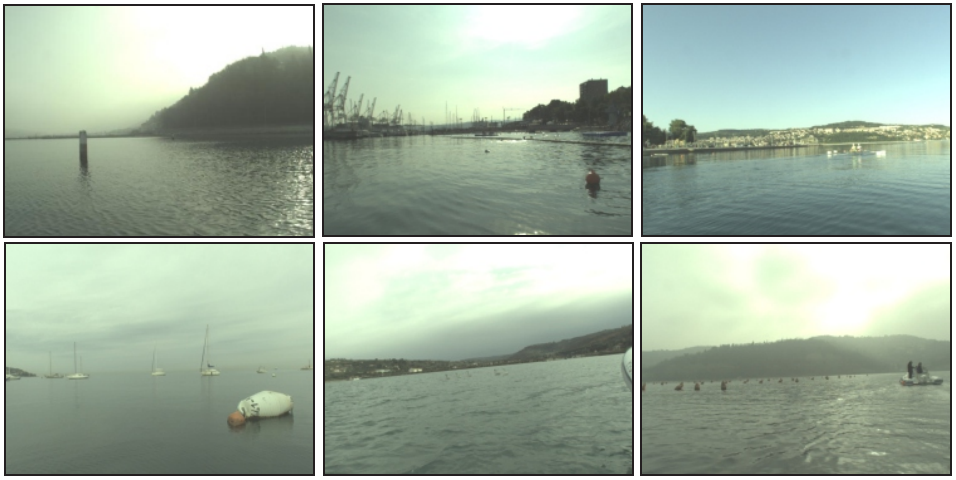}
    \caption{
    MODS covers a broad range of obstacle appearances and types. In the above examples various boats, buoys, swimmers and even swans are visible.
    }
\label{fig:modb_example_images}
\end{figure}

Approximately forty-eight hours of footage with on-board synchronized sensors (in particular, stereo cameras, IMU, compass and GPS) was captured under the described protocol. The recordings were cut into sequences with interesting navigation scenarios and out of these, $94$ sequences, jointly containing $80,828$ images were selected. In the sequence selection, care was taken to include many diverse obstacle interactions as well as phenomenons challenging for visual recognition such as prominent sun-glitter, distinct sea-foam and driving through dense shellfish farms and floating debris. Examples are shown in Figure~\ref{fig:modb_example_images}. 

\subsection{Evaluation dataset annotation protocol}

To reduce the annotation burden, while maintaining the dataset diversity, only every 10-th frame was annotated (i.e., once per second). The annotation task involved placing a tight bounding box over each dynamic obstacle and assign it a high-level label: \textit{vessel}, \textit{person} or \textit{other}. The Modd protocol from~\cite{KristanCYB2015,bovcon2018stereo} was followed for static obstacles annotation by drawing a polygon over their lower edge, where the obstacle touches the water (i.e., the water-obstacle edge). This type of annotation was chosen since the obstacle-water edge denotes the most informative part used for practical robotic navigation. For example, inaccurate segmentation of the upper part of a pier does not affect navigation, however incorrect segmentation of the part touching the water can lead to collision. Annotation examples are shown in Figure~\ref{fig:anchors} left.

The dataset was annotated in two stages. The first stage involved crowd-sourcing and in the second stage experts adjusted the annotations to guarantee high-quality results (see Figure~\ref{fig:anchors}). The two stages are described in the following.

\subsubsection{Stage 1: Crowd-sourced annotations.}

The Clickworker~\footnote{\url{https://www.clickworker.com/}} platform was used to recruit and organize the annotations, while the annotation server was setup as an Amazon AWS instance. At this stage the task was to annotate the bounding boxes over dynamic obstacles and draw polygons over the obstacle-water edges. We modified the VGG Image Annotator (VIA) v2.0.6 from~\cite{dutta2016via,dutta2019vgg} for this purpose. 

\begin{figure}
    \centering
        \includegraphics[width=0.95\columnwidth]{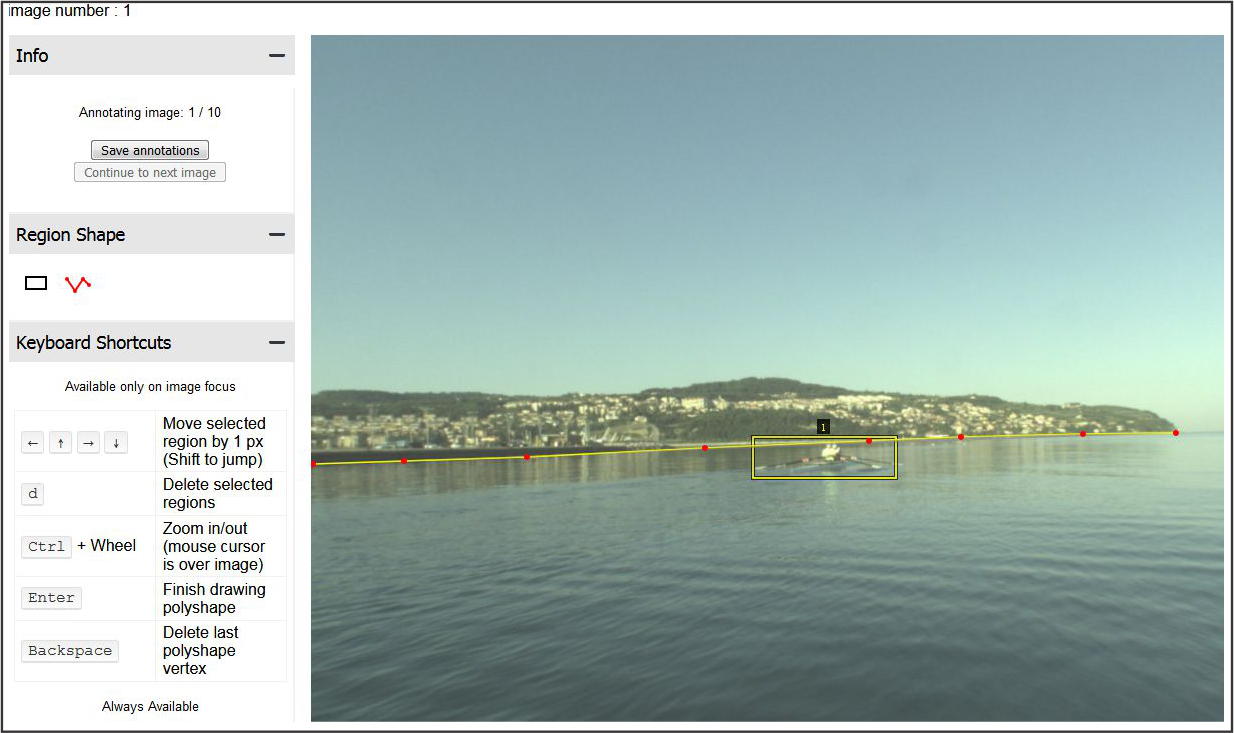}
    \caption{
    Crowd-sourcing dataset web annotation application.
    }
\label{fig:annotation_task}
\end{figure}

The work was split into tasks and presented to annotators through a web interface  (Figure~\ref{fig:annotation_task}). Each task involved annotation of 10 images, which became available only after the annotator read multiple pages of instructions with annotation illustrations. Each instruction page was followed by a short quiz to verify annotator's understanding of the task. After successful completion of instructions, a fixed password was provided to the recruits, so they could skip the quiz in future recruitment for the same task. 

The tasks were subject to two time constraints to ensure annotation quality and consistency. If an annotator finished the annotation of one image in under 60 seconds, they could not proceed to the next image before time has passed. This discouraged the annotators trying to hoard poorly executed jobs. After the batch was annotated, the system generated a unique batch code that the annotator used in the Clickworker recruitment systems to receive payment. Another time constraint was set at 3 days, during which all batches had to be finished to ensure a focused job completion, as payment codes were invalidated after that. The following steps were taken to maximize the annotation quality:
\begin{itemize}
    \item \textbf{Annotation pipeline validation}. The annotation tools and instructions were validated by assigning 100 batches of images to 100 recruits. After examining the annotation quality and average times spent on a batch, the batch fee was adjusted and additional instructions page was added to clarify the annotation requirements.
    \item \textbf{Annotator screening}. Three crowd-sourcing jobs, each containing 200 batches assigned to 200 annotators, were executed, which gave us results for 600 prospective annotators. A subset of 101 sufficiently skilled annotators was identified by careful inspection of the obtained annotations. These annotators were chosen for the final crowd-sourced annotation.
    \item \textbf{Annotation}. In consecutive 16 jobs, 50 batches of images were assigned to the skilled 101 annotators. Timed-out annotation batches were re-assigned in the subsequent jobs until all annotations were completed.
\end{itemize}

\subsubsection{Stage 2: Annotation refinement.}

\begin{figure}
    \centering
        \includegraphics[width=0.8\columnwidth]{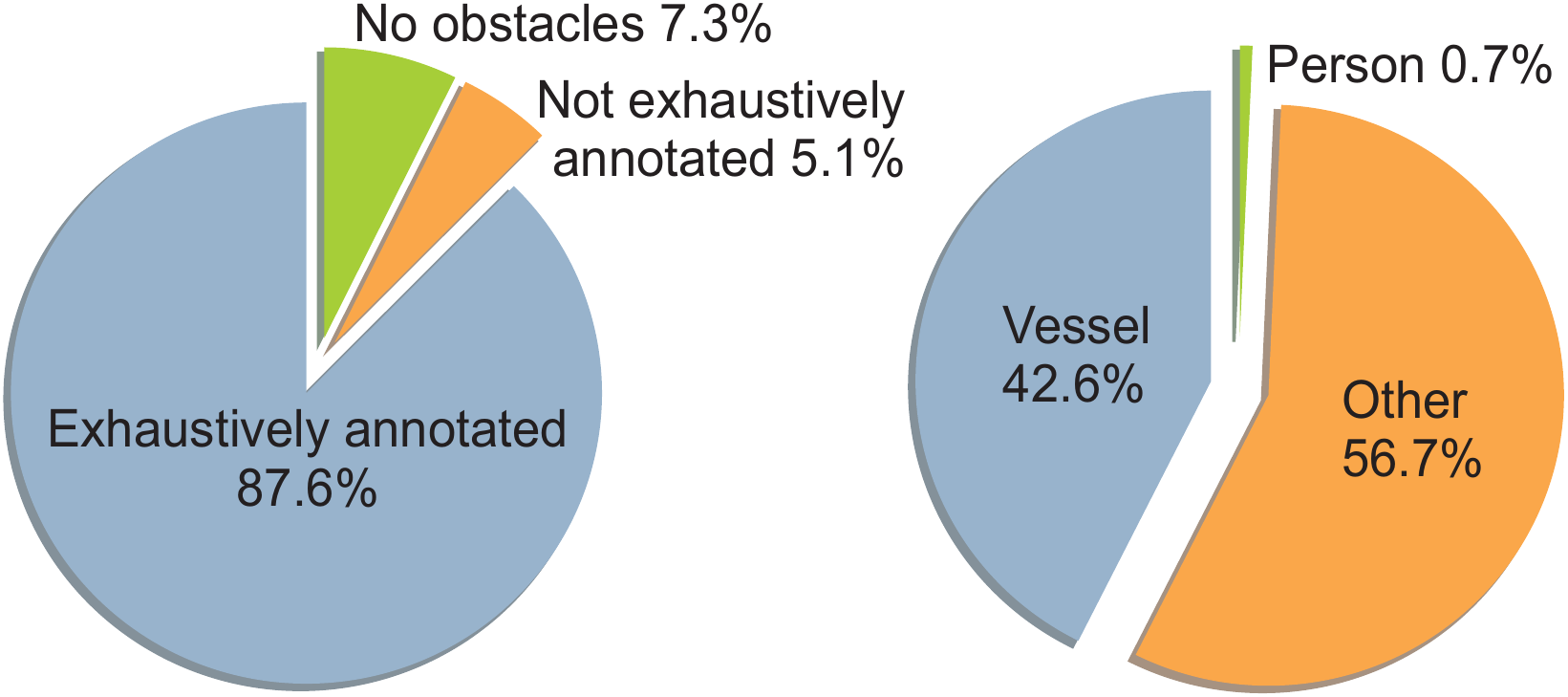}
    \caption{Proportion of exhaustively annotated frames (left) and proportion of high-level labels (right).}
    \label{fig:modb_distributions_type_frame}
\end{figure}

\begin{figure}
    \centering
        \includegraphics[width=0.95\columnwidth]{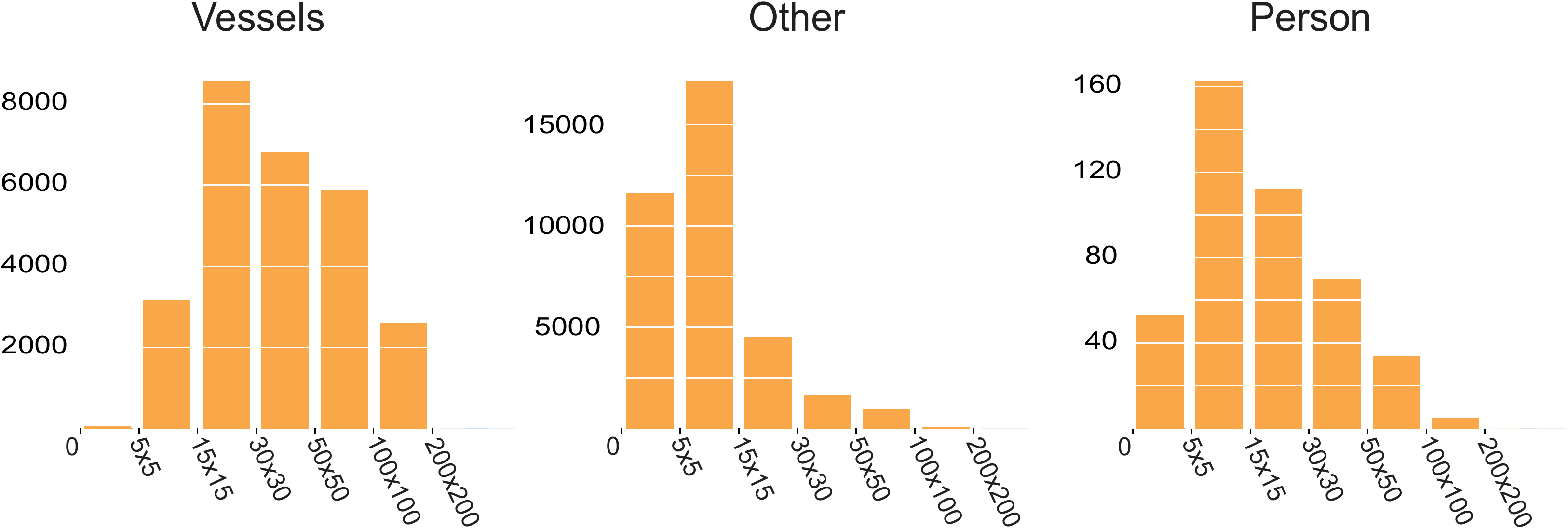}
    \caption{Distribution of obstacle sizes, grouped by their corresponding high-level label. Size is measured as surface of bounding-box in pixels.}
    \label{fig:modb_distributions_sizes}
\end{figure}

\begin{figure}
    \centering
    \includegraphics[width=0.95\columnwidth]{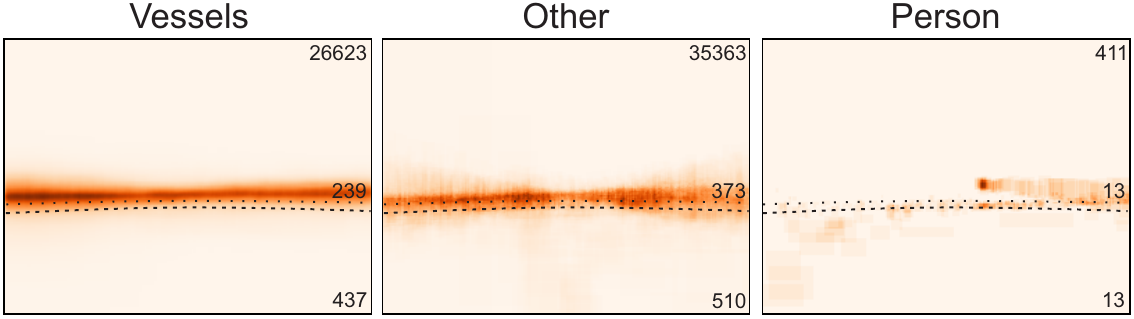}
    \caption{Heatmaps visualize frequency of instance locations in the dataset images -- darker colors indicate higher frequency. From top to bottom the numbers indicate the amount of obstacles in regions further than 30m, between 15m and 30m and closer than 15m, respectively.}
    \label{fig:heat_map}
\end{figure}

All crowd-sourced annotations were re-examined and corrected by experienced researchers with background in maritime computer vision. A Matlab tool was designed for this stage to allow easy manipulation of the existing annotations, addition of categorical labels (vessel, person, other) to the dynamic obstacles and cross-frame label propagation. The final annotations were screened by another expert to ensure labeling consistency. This amounted to $63,579$ dynamic object annotations (the proportions of high-level annotated labels is visualized in Figure~\ref{fig:modb_distributions_type_frame} right) and $10,706$ obstacle-water edge annotations which appeared in $99.3\%$ of frames. A distribution of object sizes is shown in Figure~\ref{fig:modb_distributions_sizes}, while Figure~\ref{fig:heat_map} visualizes locations frequency in the images.

Note that some images contain extemely dense object distribution and are very difficult to annotate with a high degree of accuracy (see. e.g., the shellfish farm in Figure~\ref{fig:modb_example_images}). A similar problem has been encountered in other benchmarks like COCO~\cite{lin2014coco} and most recently in LVIS~\cite{gupta2019lvis}. To keep the annotation burden manageable, we followed the approach from LVIS~\cite{gupta2019lvis} and assigned a global label to each image indicating whether all objects have been labelled. This label is then used in a federated evaluation protocol in Section~\ref{subsec:object_detection_measures} to ensure a fair use of the images with incomplete/ambiguous annotations. In some rare cases objects, like diagonally-placed mooring ropes, are poorly approximated by bounding boxes. We thus extend the LVIS approach and introduce \textit{ignore regions}, within which the performance is not evaluated.

\subsection{Training datasets}\label{subsec:training_datasets}

\begin{figure}
    \centering
        \includegraphics[width=0.95\columnwidth]{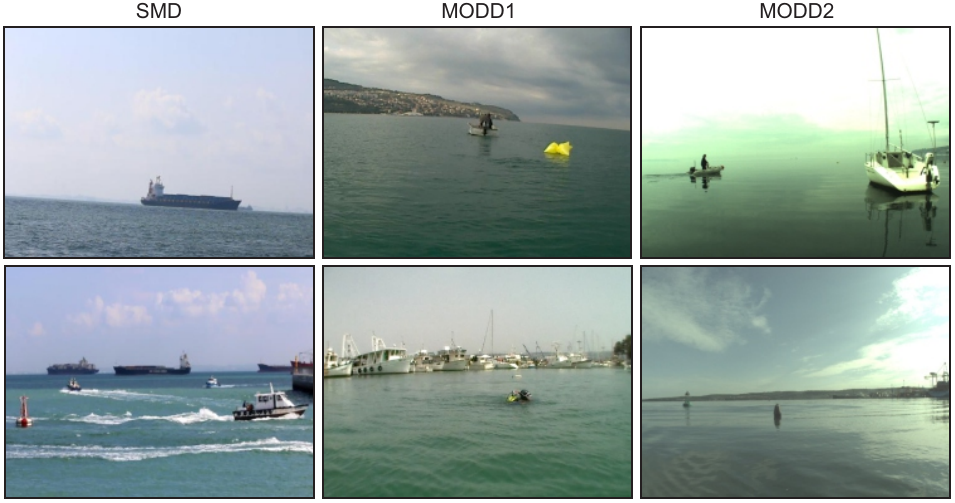}
    \caption{ Three datasets (SMD~\cite{smd_prasad2017video}, MODD1~\cite{KristanCYB2015} and MODD2~\cite{bovcon2018stereo}) with a variety of dynamic obstacles are merged and re-annotated to serve as a training dataset for maritime object detection methods.
    }
\label{fig:training_set_detector}
\end{figure}

The object detection and obstacle segmentation training datasets were collected and curated from the existing publicly available maritime datasets. Trainable object detection methods require images with annotated dynamic obstacles. We thus merged the MODD1~\cite{KristanCYB2015}, MODD2~\cite{bovcon2018stereo} and SMD~\cite{smd_prasad2017video} datasets (shown in Figure~\ref{fig:training_set_detector}) into a single MODS training dataset. The MODD1 and MODD2 are our previous USV datasets collected by the same USV as in this paper, with a slightly different acquisition hardware. The SMD dataset was added since it contains a very large collection of maritime objects recorded from shore and from a ship. All collected images were screened and images without obstacles were excluded unless they contained glitter (for hard negative mining). In addition, all images with ambiguous annotations were removed (e.g., clusters of heavily overlapping ships in marina). The annotations in the remaining images were consolidated by assigning categorical labels, manually correcting existing bounding boxes and adding missing annotations. The object detection training dataset
thus contained 24,090 images and 145,334 annotated objects.

Training of deep obstacle segmentation methods requires per-pixel labelled maritime images. We use the most recent MaSTr1325 dataset~\cite{bovcon2019mastr}, which is currently the largest maritime dataset designed specifically for training deep semantic segmentation models. The dataset has been captured by the same USV as in our evaluation dataset acquisition, albeit not in the same time period, and contains $1,325$ manually per-pixel segmented images with three semantic labels: \textit{water}, \textit{obstacle} and \textit{sky}. 

Bovcon~\etal~\cite{bovcon2019mastr} also propose dataset augmentation routines to improve generalization of the trained methods. Apart from the standard augmentation procedures, such as vertical mirroring and central rotation of $\pm \{5,15\}$ degrees, they propose using color transfer by the following procedure. In this protocol, seven images are selected from the target domain (i.e., images recorded with the same equipment as the test dataset) and the color scheme is transfered from each image to all $1,325$ reference training images. This procedure, combined with geometric augmentation, produces $158,555$ training images in the augmented dataset
. We use similar routines in training the segmentation models in MODS.

\section{Performance measures} \label{sec:performance_measures}
\subsection{Object detection by bounding boxes}
\label{subsec:object_detection_measures}

We follow the standard COCO/LVIS object detection evaluation protocol from~\cite{lin2014coco,gupta2019lvis}, which is based on the Jaccard index, i.e., an intersection-over-union (IoU) between ground truth and detected bounding boxes. A detection counts as a true positive (TP) if its respective IoU exceeds a predefined threshold, otherwise it is counted as a false positive (FP).  Because the precise localization of waterborne obstacles is difficult, especially if the objects are small, we use IoU=0.3 for the detection threshold. Precision and recall are calculated over all the images in the dataset and the F1 score is reported as the primary performance measure.

In order to focus only on the dynamic obstacles and avoid detections of people and boats on land, we use the water edge annotations to exclude detections above the water edge, unless there exists an overlap with a ground truth annotation. Thus, a detector not reporting objects above the water edge does not count as a false negative. 
As per the LVIS protocol~\cite{gupta2019lvis}, false positives are not counted in the images that are labelled as \textit{not exhaustively annotated}. 

\subsection{Obstacle detection by segmentation} \label{subsec:obstacle_detection_measures}

Segmentation methods for obstacle detection provide per-pixel labels of semantic components (water, sky, obstacles). Classical methods for segmentation evaluation, such as mean per-pixel Jaccard index~\cite{long2015fully_segmentation_metrics,cordts2016cityscapes}, do not directly reflect the accuracy meaningful for USV navigation. For example, a small patch of incorrectly segmented water pixels in the USV path should directly affect the control, while an even larger amount of incorrectly segmented pixels on top of a partially well-segmented large boat should not.

We thus follow the approach of~\cite{KristanCYB2015,bovcon2019mastr}, and evaluate separately the accuracy of obstacle-water edge estimation on static obstacles and the true-positive/false-positive measures for dynamic obstacles. Obstacle-water edge detection is analyzed in terms of localization accuracy ($\mu_\mathrm{A}$) and detection robustness ($\mu_\mathrm{R}$). The obstacle-water edge localization accuracy is defined as the square root of the average squared distance between the ground truth water edge and the per-pixel vertical nearest water edge in the segmentation mask. The obstacle-water edge detection robustness is defined as the percentage of correctly detected water edge pixels. A water-edge pixel is considered correctly detected if the vertical distance to the ground truth is below $\Theta_{\mathrm{w}}=20px$.

The dynamic obstacles detection accuracy is computed on a binary obstacle image in which obstacle pixels are set to 1 and others to 0. Care has to be taken in computation of meaningful detection accuracy since multiple objects may be fused together by the segmentation mask. Furthermore, a moderate over-segmentation should not be penalized since it will not detrimentally affect the USV navigation. 

\begin{figure}
    \centering
    \includegraphics[width=0.95\columnwidth]{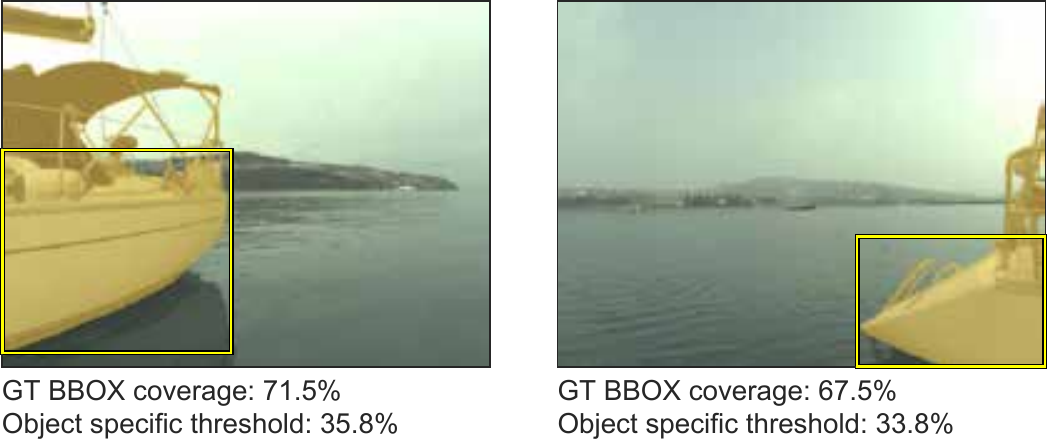}
    \caption{Illustration of object specific threshold assignment. Ground truth annotations of objects in the scene are outlined with yellow rectangles, while per-pixel semantic annotation of obstacles is displayed in orange color.}
    \label{fig:adaptive_threshold}
\end{figure}

First the true positives (TP) and false negatives (FN) are computed. A ground truth bounding box counts as a true positive if the box region contains enough pixels assigned to the obstacles in the binary obstacle mask. Otherwise, the bounding box is classified as a false negative. 
We have observed that relatively small, approximately $250$ pixels large, objects are well approximated by a bounding box -- i.e., only a fraction of pixels within the bounding box correspond to the background. But this is often not the case for larger objects in our dataset. For example, on a non-convex object, such as in Figure~\ref{fig:adaptive_threshold}, a very large portion of pixels within the ground truth bounding-box actually correspond to the background. Thus a perfect segmentation of the object might result in a false-negative detection if requiring a high overlap threshold between the segmentation mask and the bounding box. We address this problem by introducing \textit{object-specific} overlap thresholds in our dataset. For each object ground truth bounding box, we automatically determined the approximate number of pixels that correspond to the background by running an automatic segmentation algorithm DEXTR~\cite{dextr} on each bounding box. 
Thus, for the objects larger than $250$ pixels, the object-specific detection threshold is calculated as $\Theta = 0.5\frac{S_{\mathrm{D}}}{S_{\mathrm{B}}}$, where $S_{\mathrm{B}}$ is the area of the ground truth bounding box and $S_{\mathrm{D}}$ is the number of object pixels within the bounding box estimated by DEXTR. For objects smaller or equal to $250$, a fixed threshold $\Theta = 0.5$ is used.

False positives (FP) are computed next. To prevent counting false positives on slight over-segmentations of the water edge, each ground truth water edge segment is extended horizontally at both ends by $\alpha=1\%$ of image width and for each pixel, the edge location is vertically aligned with the lowest edge of the connected component the segment passes through. All pixels above the adjusted obstacle water edge are removed from the binary image and connected components are re-computed. 

In some cases, several near-by detections may be merged into a single segmentation component. To ensure that a component cannot be counted as a FN due to insufficient overlap with a ground truth bounding box and then counted as a FP in the subsequent verification, the following test is performed. If a component does not contain any assignments, it counts as a FP.

For each of the remaining components we compute the effective number of meaningful false-positives potentially merged with the detected TPs as follows. The width of the ground truth TPs assigned to the component are subtracted from the component width and the remaining value is divided by the width of the largest assigned TP scaled by a small factor $\alpha_\mathrm{sc}=1.1$. The small scaling factor is used to prevent penalizing slight over-segmentations. The obtained value rounded down to the closest integer is the effective number of FPs associated with the component. Note that, like in object detection measures, the FPs are computed only on images labelled as fully annotated.

Finally, the obstacle detection accuracy is expressed by the precision ($\mathrm{Pr} = \frac{\mathrm{TP}}{\mathrm{TP} + \mathrm{FP}}$), recall ($\mathrm{Re} = \frac{\mathrm{TP}}{\mathrm{TP} + \mathrm{FN}}$) and their harmonic mean, i.e., $\mathrm{F1} = \frac{2\cdot\mathrm{Pr}\cdot\mathrm{Re}}{\mathrm{Pr}+\mathrm{Re}}$. Additionally, we report the average number of true positive (TPr) and false positive (FPr) detections per one hundred images to provide a better insight into the quantity of triggered detections. 

\subsection{Danger zone}
\label{subsec:danger_zone}

\begin{figure}
    \centering
    \includegraphics[width=0.95\columnwidth]{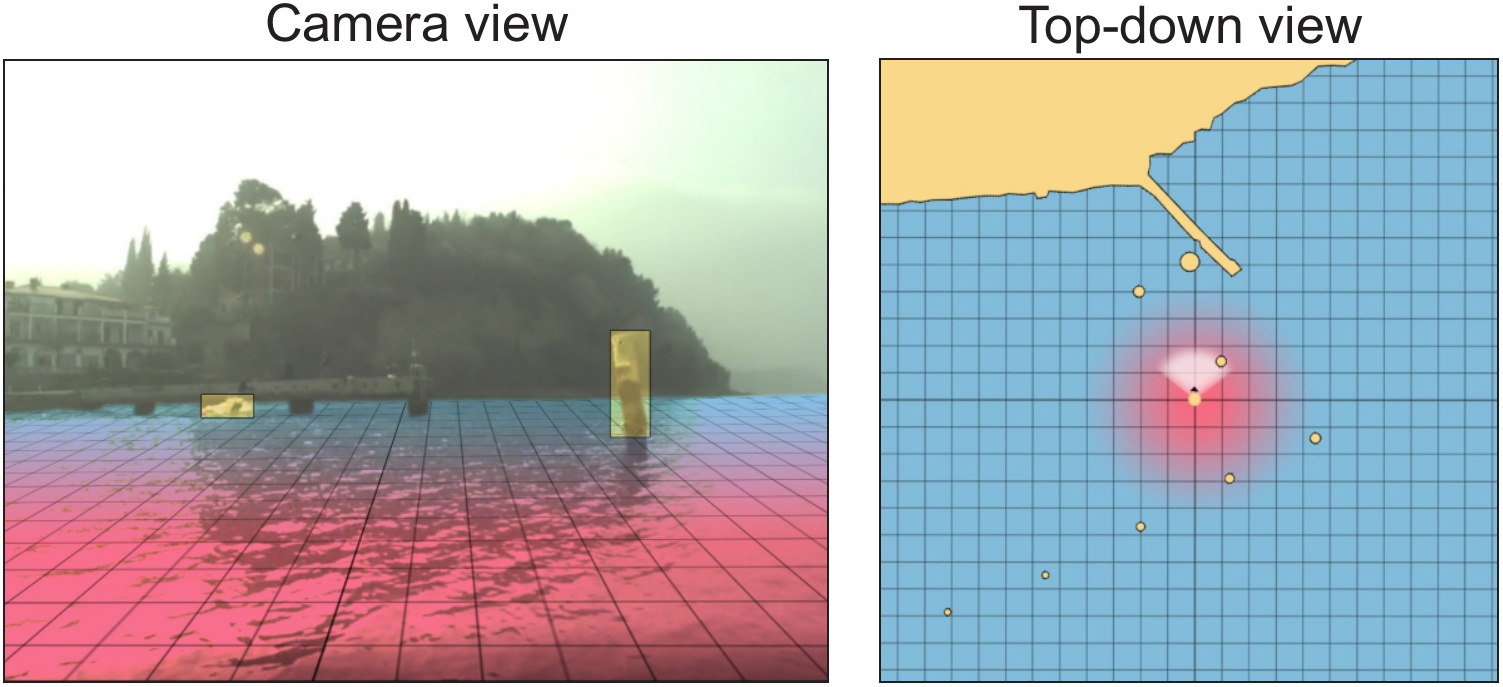}
    \caption{Nearby objects require immediate attention to avoid potential collision. A 15m hazardous area around USV (i.e., danger zone) is thus specified and visualized by a color gradient, ranging from red (dangerous) to blue (safe).}
    \label{fig:danger_zone}
\end{figure}

The danger that obstacles pose to the USV depends on their distance. Obstacles located in close proximity are more hazardous than distant ones. To address this, \cite{bovcon2019mastr} defined a danger zone as a radial area, centered at the location of the USV. The radius is chosen in such a way, that the farthest point of the area is reachable within ten seconds when travelling continuously with an average speed of $1.5$m/s. Following~\cite{bovcon2019mastr} we thus estimate the danger zone in each image (see Figure~\ref{fig:danger_zone}) from the camera-IMU geometry and report the performance both on the whole image as well as constrained only to the danger zone.

\section{The benchmark}\label{sec:benchmark}

In the following we evaluate the performance of state-of-the-art object detection methods (Section~\ref{subsec:detectors_results}) and obstacle segmentation methods (Section~\ref{subsec:segmentation_results}) on MODS. All experiments in this section were run on a desktop computer with Intel Core i7-7700 3.6GHz CPU and Nvidia RTX2080 Ti GPU with 11GB GRAM.

\subsection{Maritime object detection evaluation}
\label{sec:det_benchmark_experiments}

Several state-of-the-art general object detection methods fine-tuned to the maritime domain were considered in our experiments. MaskRCNN~\cite{he2017mask} was selected because of its good performance and widespread use in different object detection tasks. The fourth iteration of the Yolo object detector was selected because of its single shot properties and great speed~\cite{bochkovskiy2020yolov4}. A recent FCOS~\cite{tian2019fcos} was selected because of its reported state-of-the-art results and its anchor-less approach. In addition, SSD~\cite{liu2016ssd} was selected in order to test the quality of detection on smaller images and to evaluate whether fast detectors could be used for real-time obstacle detection.
Table~\ref{tab:cnn_trainable_parameters_detectors} shows the number of parameters of each architecture as well as their per-frame inference speed.

\begin{table}
    \small{
    \resizebox{\ifdim\width>\columnwidth\columnwidth\else\width\fi}{!}{%
    \begin{tabular}{lrr}
    Architecture                                    & N$_\mathrm{param}$ & $\Delta_{t}$ [ms]\\
    \midrule
    SSD~\cite{liu2016ssd}                           & 24M           & 19.7 \\
    Fcos~\cite{tian2019fcos}                        & 50M           & 69.3 \\
    Yolo4~\cite{bochkovskiy2020yolov4}              & 61M           & 40.6 \\
    MaskRCNN~\cite{he2017mask}                     & 63M           & 109.7 \\
    \end{tabular}
    }%
    }
    \caption{The number of trainable parameters (N$_\mathrm{param}$) in the tested deep object detection methods and their per-frame processing time.}
    \label{tab:cnn_trainable_parameters_detectors}
\end{table}

We used a Keras implementation of MaskRCNN and a PyTorch implementation of FCOS and SSD. For training Yolov4 we used the Darknet framework. All architectures used were pre-trained on ImageNet, then fine-tuned using the obstacle detection training dataset from Section~\ref{subsec:training_datasets} to predict the three semantic classes: \textit{vessel}, \textit{person} and \textit{other}.

\subsubsection{Results}
\label{subsec:detectors_results}
The performance of each object detector was evaluated on the MODS evaluation dataset from Section~\ref{sec:dataset_construction}. The methods were evaluated under three evaluation setups described in the following.

$Setup_1$ is the standard object detection evaluation protocols, with the exception of only using part of the image where the obstacles can appear as discussed in Section~\ref{subsec:object_detection_measures}.

$Setup_2$ follows $Setup_1$, except that object class information is ignored to evaluate dynamic obstacle detection performance, which does not require class identification, and is crucial for path planning and collision avoidance.

$Setup_3$ follows $Setup_2$, but analyzes the performance only within the USV danger zone (Section~\ref{subsec:danger_zone}). This setup analyzes detection capability important for avoiding obstacles that pose immediate collision threat to the USV.

\begin{figure}
    \centering
    \includegraphics[width=0.95\columnwidth]{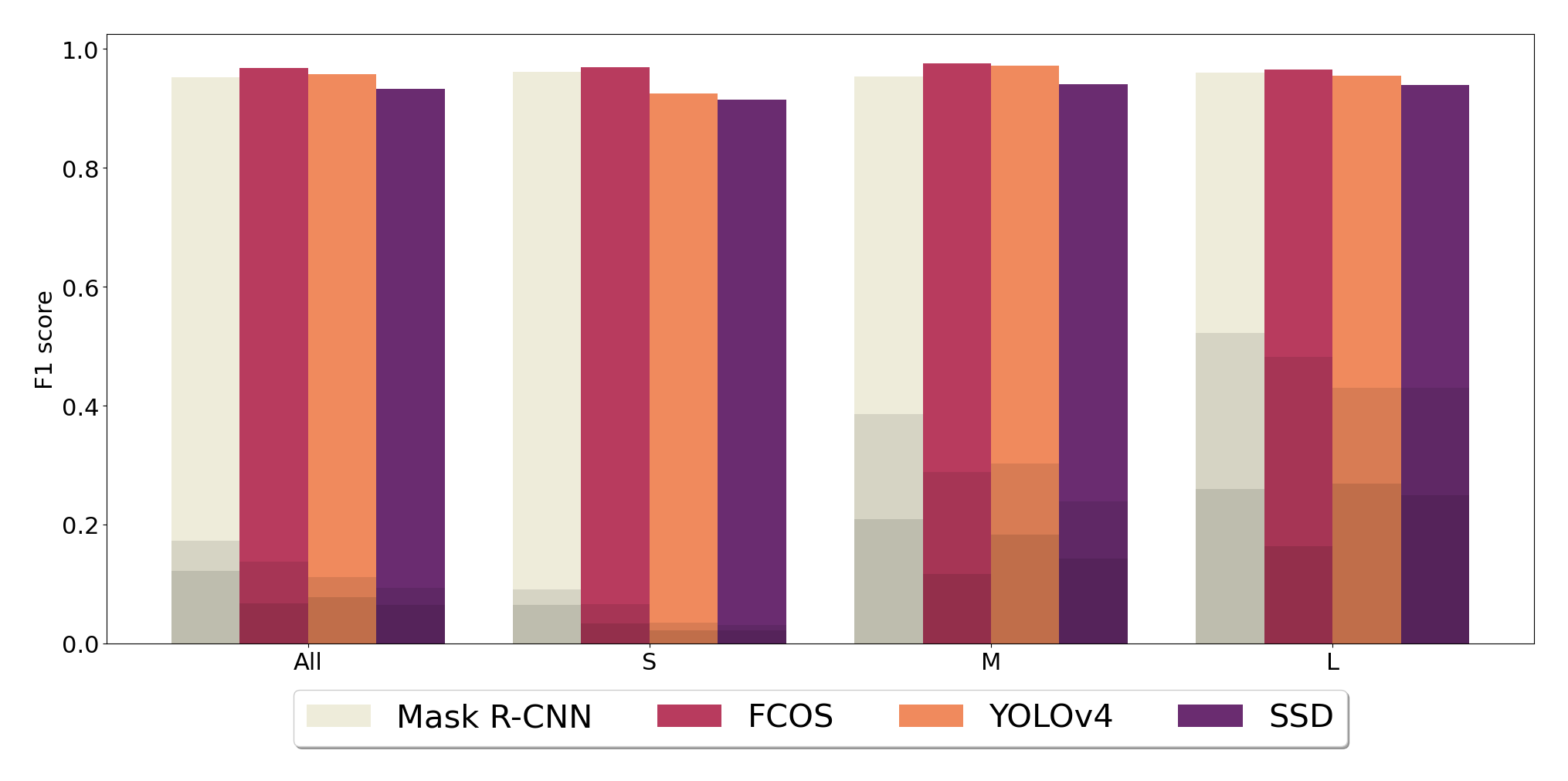}
    \caption{F1 scores for obstacles of different sizes. The darkest color in a column denotes $Setup_1$ score, a brighter color is used for $Setup_2$ scores and the brightest color depicts $Setup_3$ scores. Sizes are defined in COCO as follows: small $area<32^2$, medium $32^2 < area < 96^2$, large $area > 96^2$.}
    \label{fig:det_quantitative_comparison2}
\end{figure}

For each of the experiments, we report the overall F1 score for all annotations and F1 score for each of the size categories of the COCO evaluation protocol (small, medium and large) in order to further show the strengths and weaknesses of the evaluated detectors. The results are reported in Table~\ref{tab:modb_obstacle_detection_results} and Figure~\ref{fig:det_quantitative_comparison2}.

\begin{figure*}
    \centering
    \includegraphics[width=0.99\columnwidth]{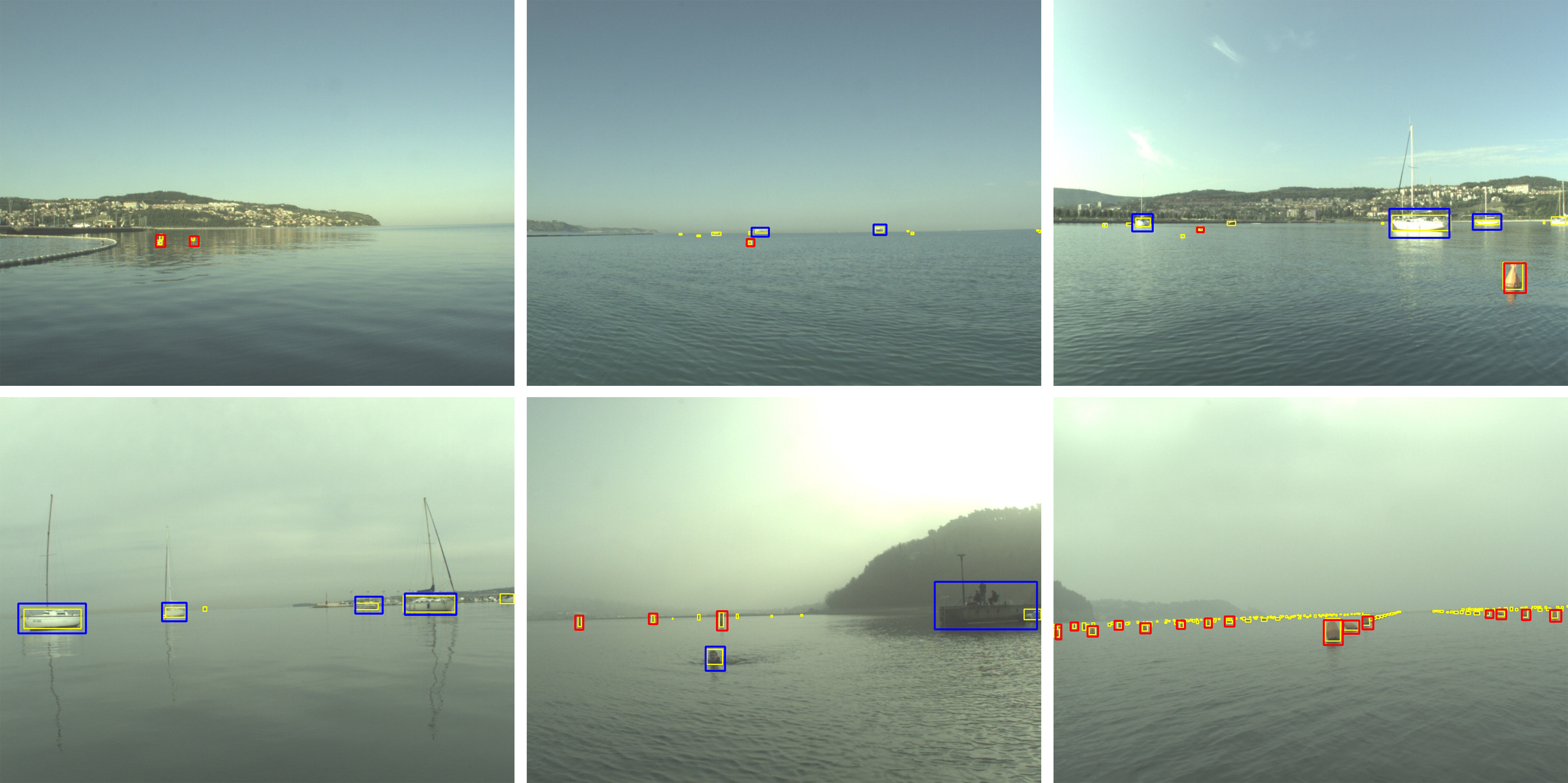}
    \caption{Examples of annotations and MaskRCNN detections from Modd3 dataset. Ground truth annotations are marked with yellow, while detections are depicted with blue, orange and red for categories \textit{vessel}, \textit{person} and \textit{other}, respectively.}
    \label{fig:det_qualitative_comparison}
\end{figure*}

Overall, the MODS images pose a significant challenge to the standard object detectors. While the precision is overall good, small objects are frequently missed. This is especially obvious when looking at $Setup_3$ scores, which lie around 0.95, while the mean scores of $Setup_1$ and $Setup_2$ are significantly lower at 0.08 and 0.12 respectively. Compared to the detection results on COCO dataset, on which state-of-the-art detectors obtain approximately 0.65 F1 score, the scores on MODS are quite low. This can be likely attributed to the specific domain with many small objects at large distances that are difficult to discriminate from waves, foam and reflections, and are also quite difficult to classify correctly.

MaskRCNN usually performs the best in $Setup_1$ and $Setup_2$ (see examples in Figure~\ref{fig:det_qualitative_comparison}), except for large objects in $Setup_1$, where it is outperformed by YOLOv4. While all detectors perform comparatively on $Setup_3$, a large variance in scores can be seen for $Setup_1$ and $Setup_2$. Specifically for small objects, MaskRCNN outperforms the other detectors by a large margin (about twice better that the second highest performing FCOS). It can also be observed that determining the correct class is quite difficult at all scales since the score difference between $Setup_1$ and $Setup_2$ is quite significant. SSD mostly performed the worst, but given the relatively small number of parameters that can be expected. However, on $Setup_3$ it performs competitively to the other methods. FCOS performed the best at all scales in $Setup_3$, despite being suboptimal in other experiments and scales.

In conclusion, all the evaluated detectors can be reliably used for detection of large obstacles within the USV danger zone. Detecting and classifying small objects at large distances, however, remains an open problem and further research is required to improve detection methods and adapt their architectures to the maritime domain.

\begin{table}
    \begin{tabular}{ccccc}
     & MaskRCNN & FCOS & YOLOv4 & SSD \\ \hline
    \textit{Setup\textsubscript{1}} & \textbf{0.125} & 0.07 & 0.08 & 0.065 \\
    \textit{Setup\textsubscript{2}} & \textbf{0.178} & 0.138 & 0.113 & 0.094 \\
    \textit{Setup\textsubscript{3}} & 0.953 & \textbf{0.969} & 0.958 & 0.933
    \end{tabular}
    \caption{F1 scores for experiments $Setup_1$, $Setup_2$ and $Setup_3$ for all the evaluated object detectors.}
    \label{tab:modb_obstacle_detection_results}
\end{table}

\subsection{Obstacle segmentation evaluation}
\label{sec:obstacle_segmentation_evaluation}
 
We evaluated sixteen recent state-of-the-art methods with reported excellent performance in various scene perception domains. Two methods, i.e., DeFP~\cite{muhovic2019obstacle} and ISSM~\cite{bovcon2018stereo} are handcrafted approaches based on traditional computer vision techniques. These two were selected since they are the most recent among state-of-the-art handcrafted maritime obstacle segmentation methods. All fourteen remaining methods are based on deep learning. PSPNet~\cite{zhao2017pyramid}, SegNet~\cite{badrinarayanan2017segnet}, RefineNet~\cite{lin2017refinenet}, DeepLab3+~\cite{chen2018deeplab3} and BiSeNet~\cite{yu2018bisenet} were chosen for their reported state-of-the-art performance on autonomous car scene segmentation and general-purpose segmentation tasks. MaskRCNN~\cite{he2017mask} and Panoptic DeepLab (DeepLab\textsubscript{pan})~\cite{cheng2020panoptic} were chosen as representatives of emerging panoptic methods that simultaneously perform object detection and instance segmentation. FuseNet~\cite{hazirbas2016fusenet} was chosen as a multimodal deep learning method that jointly considers depth and visual data for the segmentation prediction. ENet~\cite{paszke2016enet}, fully convolutional DenseNet56~\cite{jegou2017one} (denoted as FCDeNet56), fully convolutional DenseNet103~\cite{jegou2017one} (denoted as FCDeNet103), MobileUNet~\cite{howard2017mobilenets}, were selected as lightweight state-of-the-art methods that allow potentially fast processing. The last two networks,  IntCatchAI~\cite{steccanella2020intcatch} and WaSR~\cite{bovcon2020wasr_icra} are the most recent segmentation networks designed specifically for maritime environment with reportedly high obstacle detection performance.

All tested networks were adjusted to output three semantic categories (sky, water, obstacle) at each pixel. The networks backbones were initialized by pre-trained values (obtained from the published implementations). In some networks the last layer was modified to match the required three outputs. In these, the modified layer was initialized by the Xavier method~\cite{glorot2010understanding}. All deep segmentation networks were implemented in Tensorflow and trained for $10$ epochs with early-stopping enabled. We used the RMSProp optimizer~\cite{hinton2012neural} with a momentum $0.9$, initial learning rate $10^{-5}$ and standard polynomial reduction decay of $0.9$.

\begin{table}
    \small{
    \resizebox{\ifdim\width>\columnwidth\columnwidth\else\width\fi}{!}{%
    \begin{tabular}{lrr}
    Architecture                               & N$_\mathrm{param}$ & $\Delta_{t}$ [ms]\\
    \midrule
    ENet~\cite{paszke2016enet}                &  0.4M & 17.1 \\
    FCDeNet56~\cite{jegou2017one}             &  1.4M & 51.2 \\
    IntCatchAI~\cite{steccanella2020intcatch} &  1.9M &  7.1 \\
    MobileUNet~\cite{howard2017mobilenets}    &  8.9M & 31.2 \\
    FCDeNet103~\cite{jegou2017one}            &  9.3M & 80.4 \\
    SegNet~\cite{badrinarayanan2017segnet}    & 35.0M & 33.5 \\
    FuseNet~\cite{hazirbas2016fusenet}        & 44.2M & 75.1 \\
    DeepLab\textsubscript{pan}~\cite{cheng2020panoptic} & 46.7M & 120.6\\
    BiSeNet~\cite{yu2018bisenet}              & 47.5M & 17.7 \\
    DeepLab3+~\cite{chen2018deeplab3}         & 48.0M & 12.6 \\
    PSPNet~\cite{zhao2017pyramid}             & 56.0M & 22.2 \\ 
    WaSR~\cite{bovcon2020wasr_icra}           & 84.6M & 63.9 \\
    RefineNet~\cite{lin2017refinenet}         & 85.7M & 38.5 \\
    \end{tabular}
    }%
    }
    \caption{The number of trainable parameters (N$_\mathrm{param}$) in the tested obstacle segmentation methods and their average time needed to perform segmentation of a single image, measured in milliseconds.}
    \label{tab:cnn_trainable_parameters_segm}
\end{table}

\subsubsection{Results}
\label{subsec:segmentation_results}

\begin{figure}
    \centering
        \includegraphics[width=0.95\columnwidth]{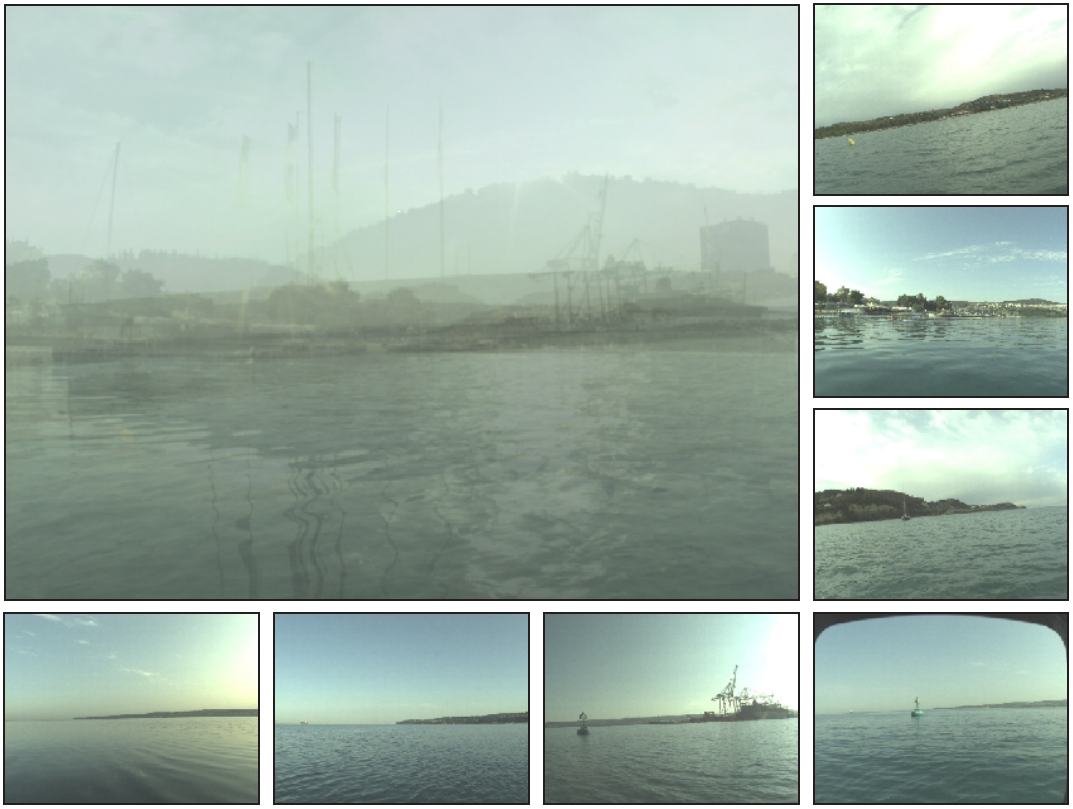}
    \caption{Seven representative images used in color-transfer technique and their mean image.}
\label{fig:images_for_augmentation}
\end{figure}

The first experiment evaluated the impact of color augmentation approach on the network performance. Following~\cite{bovcon2019mastr}, we applied geometric augmentation and color transfer technique described in Section~\ref{subsec:training_datasets} to augment and adapt the MaSTr1325 training set to our target domain -- we denote this technique as A$_{\mathrm{GC}}$. Figure~\ref{fig:images_for_augmentation} shows the seven reference images from the target domain used in the color transfer. To evaluate the impact of the dataset augmentation on the tested methods, we include two additional augmentation variations. The first variation, A$_{\mathrm{G}}$, is the same as A$_{\mathrm{GC}}$, except without color transfer, while the second variation A$_{\mathrm{GC2}}$ applies color transfer using an average of the seven images as the reference (see Figure~\ref{fig:images_for_augmentation}).

The results for the different augmentation methods are summarized in Table~\ref{tab:modb_results_obstacle_segmentation_augmentations}. Note, that the hand-crafted obstacle detection methods (ISSM and DeFP) were excluded from this experiment, considering they do not undergo the training procedure. The results indicate, that shallow networks with few parameters (MobileUNet, IntCatchAI) tend to perform better when trained with the A$_{\mathrm{GC}}$ augmentation technique. On the other hand, the deeper networks apparently over-fit to the additional color-augmented images and the performance deteriorates compared to the variants trained with the A$_{\mathrm{G}}$ and A$_{\mathrm{GC_2}}$ augmentation techniques. DeepLab\textsubscript{pan} and RefineNet, trained with the A$_{\mathrm{GC_2}}$ augmentation technique, achieve the overall best results, based on F1 measure. However, a large majority of networks, RefineNet and DeepLab\textsubscript{pan} included, perform best within the danger zone when trained with A$_{\mathrm{G}}$. Since this is the most crucial region for safe navigation, we use the A$_{\mathrm{G}}$ augmentation technique in the following experiments. 

\begin{table}
    \small{
\resizebox{\ifdim\width>\columnwidth\columnwidth\else\width\fi}{!}{%
\begin{tabular}{lccc}
Architecture      & $\mathrm{A_{G}}$ & $\mathrm{A_{GC}}$ & $\mathrm{A_{GC_2}}$ \\
\midrule
\rowcolor{l_gray}ENet       &  73.8 / 29.1                   &  69.4 / 23.6                   &  80.3 / 60.0                 \\
\rowcolor{l_gray}MobileUNet &  81.6 / 55.4                   &  83.9 / 58.8                   &  81.2 / 47.2                 \\
\rowcolor{l_gray}IntCatchAI &  57.7 / 17.3                   &  76.8 / 49.4                   &  63.2 / 15.1                 \\

FCDeNet56                   &  84.7 / 70.2                   &  81.3 / 51.6                   &  83.9 / 53.4                 \\
FCDeNet103                  &  82.7 / 66.3                   &  86.5 / 68.4                   &  86.9 / 68.7                 \\ 

\rowcolor{l_gray}FuseNet    &  77.8 / 52.8                   &  82.7 / 49.5                   &  80.7 / 46.5                 \\
\rowcolor{l_gray}PSPNet     &  78.9 / 37.9                   &  83.7 / 47.9                   &  79.3 / 35.1                 \\ 
\rowcolor{l_gray}SegNet     &  83.8 / 55.2                   &  87.6 / 76.3                   &  88.3 / 79.2                 \\ 
\rowcolor{l_gray}DeepLab3+  &  85.9 / 59.9                   &  80.7 / 60.9                   &  87.0 / 74.2                 \\
\rowcolor{l_gray}DeepLab\textsubscript{pan} & 91.2 / 79.4    &  \textbf{90.2} / \textbf{70.0} &  92.7 / 63.6                 \\
\rowcolor{l_gray}BiSeNet    &  90.3 / 83.7                   &  86.1 / 63.3                   &  88.8 / 76.4                 \\
\rowcolor{l_gray}RefineNet &  91.0 / 79.2                   &  81.7 / 68.4                   &  \textbf{92.8} / 74.3        \\ 
\rowcolor{l_gray}WaSR       &  \textbf{91.4} / \textbf{88.0} &  85.3 / 67.9                   &  90.2 / \textbf{86.0}                        
\end{tabular}
}%
}
    \caption{F1 measures (average and within danger zone) for the proposed three variations of the training set augmentation. The row shading indicates groups of methods with similar architecture.}
    \label{tab:modb_results_obstacle_segmentation_augmentations}
\end{table}

\begin{figure}
    \centering
    \includegraphics[width=0.95\columnwidth]{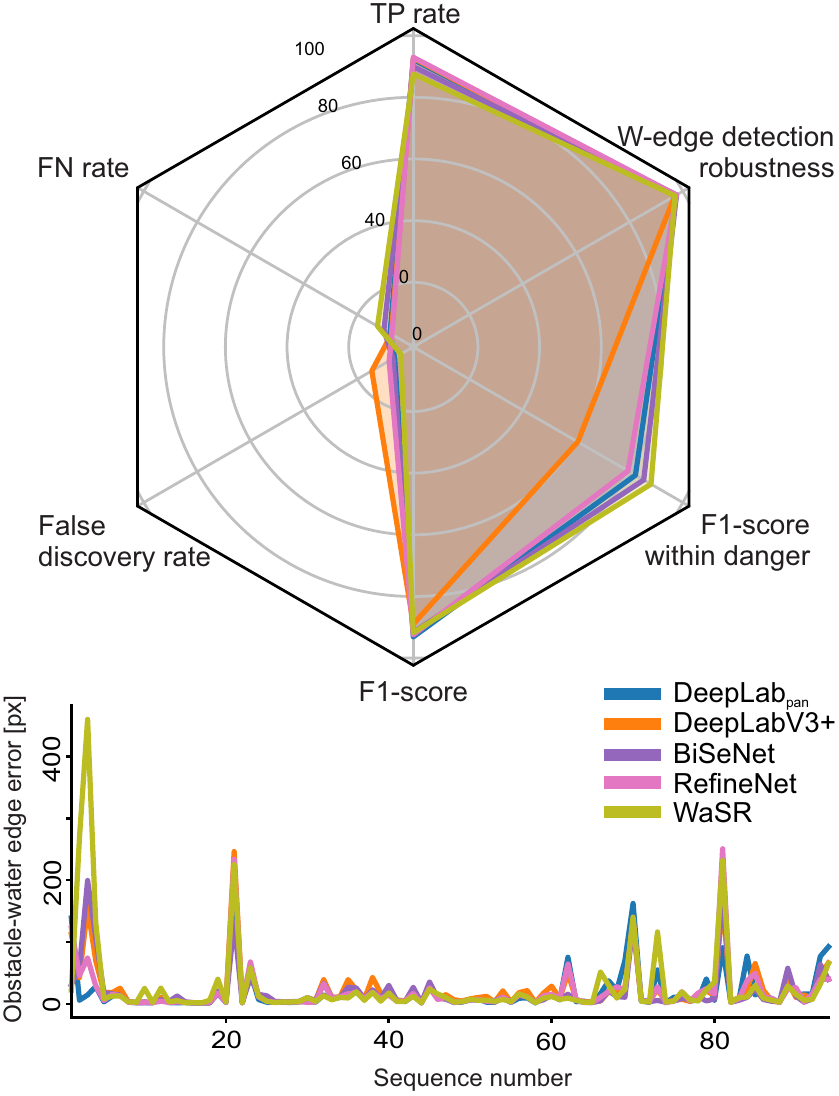}
    \caption{Quantitative comparison of the five top-performing segmentation methods in dynamic obstacle detection (radar plot) and in water edge localization error $\mu_{\mathrm{A}}$ (per-sequence graph).}
    \label{fig:segm_quantitative_comparison}
\end{figure}

\begin{figure*}
    \centering
    \includegraphics[width=0.95\columnwidth]{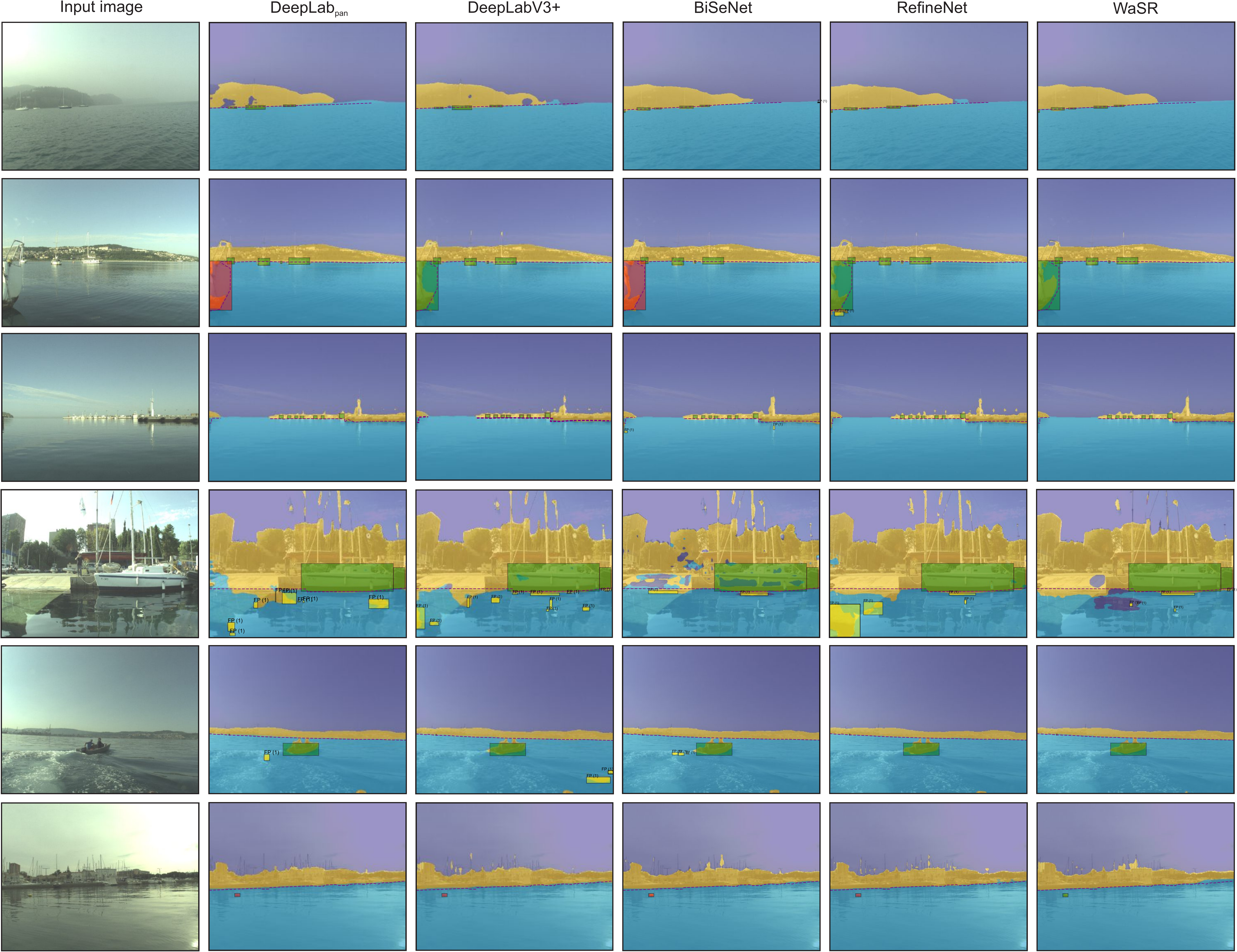}
    \caption{Qualitative comparison of the top five performing segmentation methods. Correctly detected obstacles are outlined with a green rectangle, while false detections and missed detections are marked with an orange and red rectangle, respectively. The estimated water-obstacle edge is depicted with a dotted white polygon.}
    \label{fig:segm_qualitative_comparison}
\end{figure*}

Additional insights are shed by the obstacle-water edge estimation for static obstacles and the obstacle-detection metrics, more precisely \textit{precision} and \textit{recall} combined with \textit{true-positive-rate} and \textit{false-positive-rate}, for dynamic obstacles, shown in Table~\ref{tab:modb_results_obstacle_segmentation}. The hand-crafted DeFP method was excluded from the water-edge estimation experiment, due to the limitations of the stereo camera system arising from the provided baseline distance. Similarly, the MaskRCNN method was also excluded from this experiment as it is unable to produce semantic segmentation prediction for arbitrary obstacles, such as land or piers. The ISSM method is capable of detecting the obstacle-water edge, but this is often undershoot, which in turn leads to poor overall assessment and very limited navigable area. The overall obstacle-water edge localization accuracy is best for BiSeNet, DeepLab\textsubscript{pan}, RefineNet, FCDeNet103. Among these, FCDeNet103 and RefineNet tend to more often undershoot the water-edge. Undershooting the obstacle-water edge affects the usability of the USV as it conservatively limits the navigable area, however, overshooting the obstacle-water edge leads to dangerous situations, where the USV may collide with static obstacles. On average, segmentation methods tend to overshoot the obstacle-water edge, which happens $90.5\%$ of the time when the observed edge is falsely estimated. Although WaSR utilises an additional IMU data for the water-edge refinement, its obstacle-water edge estimation accuracy is lower than some of the other deep learning methods that do not use such information. Figure~\ref{fig:segm_quantitative_comparison} shows that the error is dominated by sequences 2 and 3. Detailed inspection showed that these scenes contain a narrow band of swimming area close to shore, delimited by a floating barrier. This band of water is incorrectly segmented as an obstacle and the water edge is undershot. 
Another set of error spikes in water-obstacle edge localization accuracy, which are common to all methods, appears in sequences 21, 66, 70 and 81. In sequence 21 all observed methods undershoot the obstacle-water edge at the near-by slipway as illustrated in third row of Figure~\ref{fig:segm_qualitative_comparison}. In sequences 66, 70 and 81 all observed methods fail to segment a large part of distant land, surrounded by fog, as an obstacle (Figure~\ref{fig:segm_qualitative_comparison} first row), leading to an increase in obstacle-water edge localization error. 
BiSeNet and RefineNet achieve the best water-obstacle edge detection robustness ($97.6\%$), tightly followed by DeepLab\textsubscript{pan}, FCDeNet103, WaSR and DeepLab3+ with $97.5\%$, $97.4\%$, $97.1\%$, and $97.0\%$ detection rates, respectively.

\begin{table}
    \small{
\resizebox{\ifdim\width>\columnwidth\columnwidth\else\width\fi}{!}{%
\begin{tabular}{lcccccc}
Architecture            & $\mu_{\mathrm{A}}$[px]($\mu_{\mathrm{R}}$)      & Pr           & Re            & TPr  & FPr              & F1     \\
\midrule
\rowcolor{l_gray}
ISSM                    & 181 (68.3)                   &  55.3 & 85.1 & 55.3 & 44.6 & 67.1 \\
\rowcolor{l_gray}
DeFP                    & /                           &  33.8 & 2.7 & 1.8 & 3.4 & 5.0 \\
\rowcolor{l_gray}
MaskRCNN                & /                           & 93.1 & 41.7 & 27.1 & \textbf{2.0} & 57.6 \\

ENet                    & 78 (85.9)                   & 59.8 & \textbf{96.3} & \textbf{62.6} & 42.0 & 73.8 \\
MobileUNet              & 35 (94.6)                   & 79.1 & 84.4 & 54.8 & 14.5 & 81.6 \\
IntCatchAI              & 103 (82.9)                   & 49.8 & 67.9 & 44.1 & 44.4 & 57.5 \\

\rowcolor{l_gray}
FCDeNet56               & 21 (96.7)                   & 81.3 & 88.3 & 57.4 & 13.2 & 84.7 \\
\rowcolor{l_gray}
FCDeNet103              & 18 (97.4)                   & 75.6 & 91.4 & 59.3 & 19.1 & 82.7 \\

FuseNet                 & 46 (92.6)                   & 76.6 & 79.0 & 51.3 & 15.7 & 77.8 \\
PSPNet                  & 21 (96.9)                   & 69.4 & 91.5 & 59.4 & 26.2 & 78.9 \\ 
SegNet                  & 23 (96.8)                   & 79.4 & 88.8 & 57.7 & 15.0 & 83.8 \\
DeepLab3+               & 21 (97.0)                   & 80.0 & 92.7 & 60.2 & 15.1 & 85.9 \\
DeepLab\textsubscript{pan} & \textbf{17} (97.5)       & 90.1 & 92.2 & 59.9 & 6.6 & 91.2 \\
BiSeNet                 & \textbf{17} (\textbf{97.6}) & 90.6 & 89.9 & 58.4 & 6.1 & 90.3 \\       
RefineNet               & 18 (\textbf{97.6})          & 89.1 & 93.0 & 60.4 & 7.4 & 91.0 \\
WaSR                    & 21 (97.1)                   & \textbf{95.6} & 87.5 & 56.8 & 2.6 & \textbf{91.4} \\
\end{tabular}
}%
}
    \caption{Performance of obstacle segmentation methods in terms of water-edge localization accuracy $\mu_{\mathrm{A}}$ and detection robustness $\mu_{\mathrm{R}}$ with the percentage of correctly detected water-edge pixels in parentheses, the average number of true positive (TPr) and false positive (FPr) detections per one hundred images. We additionally report precision (Pr), recall (Re) and F1 scores in percentages. 
    The rows shading indicates groups of methods with similar architecture.}
    \label{tab:modb_results_obstacle_segmentation}
\end{table}

The handcrafted DeFP method triggers very few detections. Since its range is limited by the stereo camera system, obstacles in distance remain undetected. Similarly, floating obstacles that do not protrude through the water surface go undetected. Consequently, both precision and recall scores are very low compared to other networks. On the other hand, the handcrafted ISSM method is very sensitive and triggers many detections. In addition to a large number of TPs, it also triggers a significant amount of FPs, which is apparent from a high recall score and poor precision. Nevertheless it still surpasses light-weight deep segmentation networks like IntCatchAI in terms of F1 score. Among deep segmentation methods, the ENet method is the most sensitive, triggering the most TPs and a large amount of FPs, both contributing to an excellent recall score and a poor precision. An abundant number of detections, combined with a conservative obstacle-water edge estimation would contribute to numerous interruptions in navigation, rendering the method safe, but not useful for practical application. Same observation holds for other sensitive methods, such as IntCatchAI and PSPNet. On the other hand, FuseNet is more robust as it tends to trigger a low number of false alarms. However, its TPr is also low compared to other deep learning methods of similar complexity, which is dangerous for practical application. For safe and uninterrupted navigation an optimal ratio between the number of correct detections (TPs) and false alarms (FPs), expressed by the F1 measure, should be reached. The highest F1 measure score is achieved by WaSR, tightly followed by DeepLab\textsubscript{pan}, RefineNet, BiSeNet and DeepLab3+, which are lagging behind by $0.2\%$, $0.4\%$, $1.1\%$ and $5.5\%$, respectively. Although WaSR is the least sensitive (Table~\ref{tab:modb_results_obstacle_segmentation} and Figure~\ref{fig:segm_quantitative_comparison} top), triggering on average merely 2.6 false alarms per one-hundred images, it also manages to detect a large number of TPs, which contributes towards its best F1 measure score. Qualitative examples in Figure~\ref{fig:segm_qualitative_comparison} further confirm, that WaSR is the most robust to the sea foam (fourth row), environmental reflections (second and third row) and other visual ambiguities (first row). Furthermore, it also correctly detects difficult obstacles within the environmental reflections (Figure~\ref{fig:segm_qualitative_comparison} last row).     

Timely detection and localization of obstacles is vital for safe autonomous navigation. Obstacles that require immediate action are located in the danger-zone area (Figure~\ref{fig:danger_zone}). The obstacle detection results from within this area are shown in Table~\ref{tab:modb_results_obstacle_segmentation_dangerzone}. Within the danger-zone, the handcrafted DeFP and ISSM methods outperform some lightweight networks (e.g. ENet) and their results are also comparable to deeper networks such as PSPNet. Most of deeper networks achieve a very high recall score with a similar true-positive-rate. However, their precision varies significantly as well as the average number of FP detections. MaskRCNN achieves both high true-positive-rate and a low false-positive rate. This is due to the fact that common maritime obstacles, such as buoys and boats, appear in the danger zone to a large extent, which object detectors can successfully detect. WaSR achieves the best compromise between false alarms and correct detections by outperforming MaskRCNN both in precision and recall, which is also reflected in F1 score and is therefore considered as the safest and the most optimal to use for the autonomous navigation in a marine environment among the tested methods.


\begin{table}
    \small{
\resizebox{\ifdim\width>\columnwidth\columnwidth\else\width\fi}{!}{%
\begin{tabular}{lcccccc}
Architecture         & Pr            & Re            & TPr          & FPr          & F1   \\
\midrule
\rowcolor{l_gray}
ISSM                 & 21.2          & 67.4          & 2.7          & 10.0         & 32.2 \\
\rowcolor{l_gray}
DeFP                 & 24.8          & 25.9          & 1.0          & 3.2          & 25.4 \\
\rowcolor{l_gray}
MaskRCNN             & 64.0          & 89.0          & 0.8          & \textbf{0.4} & 74.5 \\

ENet                 & 17.2          & 96.0          & 3.8          & 18.6         & 29.1 \\
MobileUNet           & 39.7          & 91.8          & 3.7          & 5.6          & 55.4 \\
IntCatchAI           & 9.7           & 77.2          & 3.1          & 28.7         & 17.3 \\

\rowcolor{l_gray}
FCDeNet56            & 56.5          & 92.7          & 3.7          & 2.9          & 70.2 \\
\rowcolor{l_gray}
FCDeNet103           & 51.1          & 94.2          & 3.8          & 3.6          & 66.3 \\

FuseNet              & 41.3          & 73.2          & 0.6          & 0.9          & 52.8 \\
PSPNet               & 23.6          & 95.5          & 3.8          & 12.4         & 37.9 \\
SegNet               & 39.0          & 94.3          & 3.8          & 5.9          & 55.2 \\
DeepLab3+            & 43.5          & 95.8          & 3.8          & 5.0          & 59.9 \\
DeepLab\textsubscript{pan} & 67.5    & 96.2          & \textbf{3.9} & 1.9          & 79.4 \\
BiSeNet              & 74.8          & 94.8          & 3.8          & 1.3          & 83.7 \\
RefineNet            & 67.3          & \textbf{96.3} & \textbf{3.9} & 1.9          & 79.2 \\
WaSR                 & \textbf{84.2} & 92.1          & 3.7          & 0.7          & \textbf{88.0} \\
\end{tabular}
}%
}

    \caption{Performance of obstacle segmentation methods within danger-zone in terms of precision (Pr), recall (Re) and F1 scores measured in percentages. Additionally, we report the average number of true positive (TPr) and false positive (FPr) detections per one hundred images.}
    \label{tab:modb_results_obstacle_segmentation_dangerzone}
\end{table}

A correct detection of different types of obstacles is not equally important. For example, a collision with a typical buoy may only prevent a passage, while a collision with a vessel might damage the USV and harm the potential on-board passengers. Even more alarming are collisions with swimmers which may result in a serious accident. The results from Table~\ref{tab:modb_results_obstacle_segmentation_typebased} indicate that Deeplab3+ most often correctly detects persons and vessels (within and outside the danger-zone). However, this method is very sensitive and produces twice as much FPs (Table~\ref{tab:modb_results_obstacle_segmentation} and~\ref{tab:modb_results_obstacle_segmentation_dangerzone}) as the other methods, therefore we  exclude it from further analysis. The remaining four methods (Deeplab\textsubscript{pan}, BiSeNet, RefineNet and WaSR) are approximately equally successful in detecting vessels, however RefineNet has a slight advantage both within and outside of the danger-zone. Deeplab\textsubscript{pan} performs overall the best in person detection, while RefineNet outperforms the compared methods in the other obstacle types.

\begin{table}
    \small{
\resizebox{\ifdim\width>\columnwidth\columnwidth\else\width\fi}{!}{%
\begin{tabular}{lccccc}
  & DLP                       & DLV3                      & BiSeNet        & RefineNet                 & WaSR    \\
\hline
P & 89          / \textbf{98} & \textbf{93} / 96          & 90 / 96        & 91 / 94                   & 91 / 88 \\
V & \textbf{98} / 98          & \textbf{98} / \textbf{99} & 97 / 98        & \textbf{98} / \textbf{99} & 97 / 97 \\
O & 86          / \textbf{95} & \textbf{87} / 93          & 82 / 92        & \textbf{87} / 94          & 78 / 88 \\
\end{tabular}
}%
}
    \caption{Obstacle detection rates (average and within danger zone) for five top performning methods with respect to obstacle types: persons (P), vessels (V) and other (O). DeepLab3+ and DeepLab\textsubscript{pan} are abbreviated to  DLV3 and DLP, respectively.
    }
    \label{tab:modb_results_obstacle_segmentation_typebased}
\end{table}

To analyse the obstacle detection accuracy based on their size, we consider six size classes -- tiny (less than $255$px), very small (less than $900$px), small (less than $2,500$px), medium (less than $10,000$px), large (less than $40,000$px) and very large (more than $40,000$px). The detection results are shown in Table~\ref{tab:modb_results_obstacle_segmentation_sizes_combined}. Obstacles with larger surface area are in general more accurately detected, while tiny obstacles are often missed. Most false-positive detections are medium sized, with their surface area ranging between $2,500$ and $10,000$ pixels. RefineNet achieves the best true-positive-rate (TPR) for multiple size classes. In terms of false-discovery-rate (FDR) WaSR outperforms the other methods by a large margin in each size class.

Obstacle detection performance within the danger-zone, grouped by sizes, is shown in Table~\ref{tab:modb_results_obstacle_segmentation_sizes_combined}. Most obstacles in this area are medium size or larger, with a very low percentage of smaller classes -- $18$\% (see Table~\ref{tab:modb_results_obstacle_segmentation_sizes_ratio}). Almost no obstacles within the danger-zone belong to the tiny size class. Nevertheless, this size class represents a majority of false alarms ($450$ on average over all methods). In addition, approximately $32$\% of all false alarms of this size occur within the danger zone. Medium sized and larger obstacles, on the other hand, are accurately detected with an average detection rate above $98$\% overall and within the danger zone. False alarms larger than $10,000$px are rarely triggered within the danger zone.

\begin{table*}
    \small{
\resizebox{\ifdim\width>\columnwidth\columnwidth\else\width\fi}{!}{%
\begin{tabular}{lcccccccccccc}
\multirow{2}{*}{Architecture} & \multicolumn{2}{c}{Tiny}                                  & \multicolumn{2}{c}{Very small}                            & \multicolumn{2}{c}{Small}                                 & \multicolumn{2}{c}{Medium}                                & \multicolumn{2}{c}{Large}                                 & \multicolumn{2}{c}{Very large}                            \\
                              & $\mathcal{R}_{\mathrm{TP}}$ & $\mathcal{R}_{\mathrm{FP}}$ & $\mathcal{R}_{\mathrm{TP}}$ & $\mathcal{R}_{\mathrm{FP}}$ & $\mathcal{R}_{\mathrm{TP}}$ & $\mathcal{R}_{\mathrm{FP}}$ & $\mathcal{R}_{\mathrm{TP}}$ & $\mathcal{R}_{\mathrm{FP}}$ & $\mathcal{R}_{\mathrm{TP}}$ & $\mathcal{R}_{\mathrm{FP}}$ & $\mathcal{R}_{\mathrm{TP}}$ & $\mathcal{R}_{\mathrm{FP}}$ \\
\hline
DeepLab\textsubscript{pan}  & 83.1 / \textbf{74.1}           & 8.8 / 83.3          & 98.1 / \textbf{91.9}          & 7.6 / 50.2          & 98.6 / 96.0          & 4.6 / 18.8          & 98.4 / 98.5          & 16.8 / 9.0          & 99.2 / 98.5          & 26.2 / 30.8         & 99.8 / 99.7           & \textbf{0.0} / \textbf{0.0}   \\

DeepLab3+  & \textbf{84.1} / 64.7  & 15.2 / 92.1          & 97.9 / 89.2 & 21.0 / 79.5          & 99.1 / \textbf{97.4} & 25.2 / 59.6          & 98.8 / 99.0 & 27.0 / 33.5          & 99.4 / 98.9 & 7.2 / 13.9         & 99.5 / 99.5           & 0.2 / 0.3            \\

BiSeNet    & 77.7 / 64.7           & 9.4 / 80.4          & 97.2 / 84.6          & 9.4 / 46.0          & 98.9 / 95.1          & 7.7 / 17.5          & 98.6 / 98.5          & 12.4 / 6.0 & 99.5 / 99.1          & 8.5 / 16.4         & \textbf{100.0} / \textbf{100.0}           & \textbf{0.0} / \textbf{0.0}   \\

RefineNet  & \textbf{84.1} / 66.5  & 11.5 / 86.0          & \textbf{98.4} / 91.1         & 11.3 / 60.0          & \textbf{99.2} / \textbf{97.4}          & 12.1 / 27.6          & \textbf{99.2} / \textbf{99.2}          & 8.8 / 8.4          & \textbf{99.8} / \textbf{99.4}          & 6.8 / \textbf{1.1} & 99.8 / 99.7           & 0.2 / 0.3            \\

WaSR       & 73.1 / 56.5           & \textbf{6.3} / \textbf{72.3} & 95.9 / 74.4          & \textbf{6.0} / \textbf{37.1} & 98.1 / 93.0          & \textbf{2.1} / \textbf{13.7} & 98.1 / 97.5          & \textbf{0.9} / \textbf{4.6}          & 98.8 / 97.0          & \textbf{0.5} / 1.5          & \textbf{100.0} / \textbf{100.0} & \textbf{0.0} / \textbf{0.0}   \\
\end{tabular}
}
}
    \caption{True-Positive and False-Discovery rates (average and within danger zone), denoted as $\mathcal{R}_{\mathrm{TP}}$ and $\mathcal{R}_{\mathrm{FP}}$, for the five top performing methods are reported for different size classes of obstacles within the danger-zone.}
    \label{tab:modb_results_obstacle_segmentation_sizes_combined}
\end{table*}

\begin{table*}
    \small{
\resizebox{\ifdim\width>\columnwidth\columnwidth\else\width\fi}{!}{%
\begin{tabular}{lcccccccccccc}
\multirow{2}{*}{Architecture} & \multicolumn{2}{c}{Tiny}                                  & \multicolumn{2}{c}{Very small}                            & \multicolumn{2}{c}{Small}                                 & \multicolumn{2}{c}{Medium}                                & \multicolumn{2}{c}{Large}                                 & \multicolumn{2}{c}{Very large}                            \\
                              & $\mathcal{r}_{\mathrm{T}}$ & $\mathcal{r}_{\mathrm{F}}$ & $\mathcal{r}_{\mathrm{T}}$ & $\mathcal{r}_{\mathrm{F}}$ & $\mathcal{r}_{\mathrm{T}}$ & $\mathcal{r}_{\mathrm{F}}$ & $\mathcal{r}_{\mathrm{T}}$ & $\mathcal{r}_{\mathrm{F}}$ & $\mathcal{r}_{\mathrm{T}}$ & $\mathcal{r}_{\mathrm{F}}$ & $\mathcal{r}_{\mathrm{T}}$ & $\mathcal{r}_{\mathrm{F}}$ \\
\hline
DeepLab\textsubscript{pan}  & 0.8            & 37.5          & 2.8           & 32.4          & 6.6           & 31.2          & 15.6           & 7.7           & 29.4           & 36.6          & 89.3           & /            \\

DeepLab3+ & 0.8            & 40.6          & 2.8           & 37.1          & 6.6           & 28.5          & 15.6           & 21.2          & 29.4           & 61.3         & 89.3           & 100.0            \\

BiSeNet    & 0.8            & 26.6          & 2.8           & 20.1          & 6.6           & 16.1          & 15.6           & 7.1           & 29.4           & 62.0         & 89.3           & /            \\

RefineNet  & 0.8            & 30.0          & 2.8           & 30.5          & 6.6           & 18.0          & 15.6           & 14.8          & 29.4           & 4.2          & 89.3           & 100.0            \\

WaSR       & 0.8            & 24.2          & 2.8           & 20.1          & 6.6           & 47.2          & 15.6           & 83.3          & 29.4           & 90.9        & 89.3           & /            \\
\end{tabular}
}
}
    \caption{Detection ratios of the top five performing methods within the danger zone in relation to the full screen. The True-Positive detection ratio is denoted as $\mathcal{r}_{\mathrm{T}}$, while the False-Discovery detection ratio is marked as $\mathcal{r}_{\mathrm{F}}$.}
    \label{tab:modb_results_obstacle_segmentation_sizes_ratio}
\end{table*}

The processing speed of the tested methods is presented in Table~\ref{tab:cnn_trainable_parameters_segm}. Hand-crafted methods, ISSM and DeFP, are the only ones that run on a single desktop CPU where they perform at $34.4$ms and $9.0$ms per frame, respectively. Although their average frame rate is comparable with most deep segmentation methods, their obstacle detection performance is significantly lower. Light-weight deep segmentation methods with few parameters tend to perform faster than their heavy-weight counterparts. The two exceptions are the DenseNet variations (FCDeNet56 and FCDeNet103), which are computationally demanding. IntCatchAI is the fastest, but its detection accuracy is low. Methods that produce accurate segmentation masks, required for the autonomous navigation, are all based on heavy-weight models (WaSR, RefineNet, BiSeNet, DeepLab3+, DeepLab\textsubscript{pan} and FCDeNet56) and produce moderately low frame rates (38 FPS on average). Nevertheless, their performance can still be considered real-time (with an exception of DeepLab\textsubscript{pan}), given that typical USV cameras (including the one used in dataset acquisition) run at 10Hz.

\section{Conclusion}
\label{sec:conclusion}

A new obstacle detection benchmark (MODS), which considers two major USV perception tasks, a maritime object detection and segmentation-based obstacle detection, was presented. A new dataset with carefully annotated 80,828 stereo images synchronized by IMU was created along with specified object detection and obstacle segmentation training sets. Evaluation protocols were proposed as well for objective evaluation on the considered perception tasks. Several state-of-the-art methods were analyzed using the proposed evaluation protocols.

State-of-the-art object detection methods obtain approximately six times lower F1 scores compared to their performance on general-purpose datasets like COCO, which speaks of challenges specific for the maritime domain. Detailed analysis showed that the overall precision is good, but small objects are typically missed by all methods, leading to a low recall.
This is in contrast to the current published results on maritime domain, which are constrained to evaluation on large objects. The presented dataset, however, is much more challenging and reveals the drawbacks of the general-purpose object detection methods applied to the USV setup.
The best overall detection with recognition is obtained by MaskRCNN, while FCOS outperforms all other methods in detection-only task (class labels ignored) within immediate neighborhood of the USV (danger zone), and runs nearly twice as fast. On a high-end GPU this results in approximately 14fps, which is near real-time, considering the on-board cameras frame-rate is 10Hz. 

Several general-purpose as well as maritime-specific segmentation methods were evaluated on the obstacle segmentation task.
Results show that geometric augmentation plays a more important role than the color transfer in training. In dynamic obstacle detection, the recent maritime-specific method WaSR achieves the best F1 score, which is primarily due to a very low false-positive rate and a high true-positive rate. For example, the false-positive rate is reduced by half when considering the danger zone only compared to the second-best BiSeNet, which is a general-purpose segmentation method preforming reportedly well on segmentation for autonomous cars. Most of the false-positives of best methods within the danger zone are tiny in size, which suggests performance could be improved by addressing this size category. Indeed, large objects are detected well (above 98\%), while performance drops for small objects like persons and buoys. The best methods (RefineNet and BiSeNet) detect the water-obstacle edge fairly robustly (above $97\%$), but in some cases overshoot the water edge, which increases the danger of collision. These methods perform at approximately 14fps on a high-end GPU, which may be considered near real-time.

Our results reveal that further development of maritime-specific architectures is required to make the detection methods useful for practical USV operation. More research should be invested in designing object detection methods that cope better with visually small objects to improve detection of distant and small near-by dynamic obstacles. Segmentation methods should be improved to better estimate the water-obstacle edges and reduce the false positive rate on tiny objects. Perhaps considering sequences of images could address this to some extent by either explicitly tracking detections or by designing architectures that directly operate on consecutive frames. Effort should be put in developing architectures fast enough to be run off embedded and low-power GPU, as well. We hope the presented benchmark and the obtained insights will facilitate these venues of future research, contributing towards maritime-specific safe visual perception systems.

\section*{Acknowledgment}
\noindent{This work was supported in part by the Slovenian research agency (ARRS) programmes P2-0214 and P2-0095, and the Slovenian research agency (ARRS) research project DaViMaR J2-2506. We would additionally like to thank Jure Nendl for assisting us in network evaluation process.}

\bibliographystyle{IEEEtran}
\bibliography{bibliography}

\begin{thebibliography}{10}
\providecommand{\url}[1]{#1}
\csname url@samestyle\endcsname
\providecommand{\newblock}{\relax}
\providecommand{\bibinfo}[2]{#2}
\providecommand{\BIBentrySTDinterwordspacing}{\spaceskip=0pt\relax}
\providecommand{\BIBentryALTinterwordstretchfactor}{4}
\providecommand{\BIBentryALTinterwordspacing}{\spaceskip=\fontdimen2\font plus
\BIBentryALTinterwordstretchfactor\fontdimen3\font minus
  \fontdimen4\font\relax}
\providecommand{\BIBforeignlanguage}[2]{{%
\expandafter\ifx\csname l@#1\endcsname\relax
\typeout{** WARNING: IEEEtran.bst: No hyphenation pattern has been}%
\typeout{** loaded for the language `#1'. Using the pattern for}%
\typeout{** the default language instead.}%
\else
\language=\csname l@#1\endcsname
\fi
#2}}
\providecommand{\BIBdecl}{\relax}
\BIBdecl

\bibitem{Onunka2010}
C.~Onunka and G.~Bright, ``Autonomous marine craft navigation: On the study of
  radar obstacle detection,'' in \emph{2010 11th International Conference on
  Control Automation Robotics \& Vision}, Dec 2010, pp. 567--572.

\bibitem{Jimenez2009}
A.~R.~J. Ruiz and F.~S. Granja, ``A short-range ship navigation system based on
  ladar imaging and target tracking for improved safety and efficiency,''
  \emph{IEEE Transactions on Intelligent Transportation Systems}, vol.~10,
  no.~1, pp. 186--197, March 2009.

\bibitem{Heidarsson2011}
H.~K. Heidarsson and G.~S. Sukhatme, ``Obstacle detection and avoidance for an
  autonomous surface vehicle using a profiling sonar,'' in \emph{2011 IEEE
  International Conference on Robotics and Automation}, May 2011, pp. 731--736.

\bibitem{Muresan2019}
M.~P. Muresan and S.~Nedevschi, ``Multi-object tracking of 3d cuboids using
  aggregated features,'' in \emph{2019 IEEE 15th International Conference on
  Intelligent Computer Communication and Processing (ICCP)}, 2019, pp. 11--18.

\bibitem{Kim2017}
H.~Kim, J.~Cho, D.~Kim, and K.~Huh, ``Intervention minimized semi-autonomous
  control using decoupled model predictive control,'' in \emph{2017 IEEE
  Intelligent Vehicles Symposium (IV)}, 2017, pp. 618--623.

\bibitem{Arikan2018}
A.~Arıkan, A.~Kayaduman, S.~Polat, Y.~Simsek, I.~C. Dikmen, H.~G. Bakır,
  T.~Karadag, and T.~Abbasov, ``Control method simulation and application for
  autonomous vehicles,'' in \emph{2018 International Conference on Artificial
  Intelligence and Data Processing (IDAP)}, 2018, pp. 1--4.

\bibitem{prasad2018overview}
D.~K. Prasad, C.~K. Prasath, D.~Rajan, L.~Rachmawati, E.~Rajabally, and
  C.~Quek, ``Object detection in a maritime environment: Performance evaluation
  of background subtraction methods,'' \emph{IEEE Transactions on Intelligent
  Transportation Systems}, pp. 1--16, 2018.

\bibitem{jorge2019survey}
V.~A. Jorge, R.~Granada, R.~G. Maidana, D.~A. Jurak, G.~Heck, A.~P. Negreiros,
  D.~H. dos Santos, L.~M. Gon{\c{c}}alves, and A.~M. Amory, ``A survey on
  unmanned surface vehicles for disaster robotics: Main challenges and
  directions,'' \emph{Sensors}, vol.~19, no.~3, p. 702, 2019.

\bibitem{zhang2020review}
W.~Zhang, C.-f. Yang, F.~Jiang, X.-z. Gao, and K.~Yang, ``A review of research
  on light visual perception of unmanned surface vehicles,'' in \emph{Journal
  of Physics: Conference Series}, vol. 1606, no.~1.\hskip 1em plus 0.5em minus
  0.4em\relax IOP Publishing, 2020, p. 012022.

\bibitem{albrecht2011visual_saliency1}
T.~Albrecht, G.~A. West, T.~Tan, and T.~Ly, ``Visual maritime attention using
  multiple low-level features and naive bayes classification,'' in \emph{2011
  International Conference on Digital Image Computing: Techniques and
  Applications}.\hskip 1em plus 0.5em minus 0.4em\relax IEEE, 2011, pp.
  243--249.

\bibitem{liu2008omni}
H.~Liu, O.~Javed, G.~Taylor, X.~Cao, and N.~Haering, ``Omni-directional
  surveillance for unmanned water vehicles,'' in \emph{The Eighth International
  Workshop on Visual Surveillance-VS2008}, 2008.

\bibitem{makantasis2013vision_saliency3}
K.~Makantasis, A.~Doulamis, and N.~Doulamis, ``Vision-based maritime
  surveillance system using fused visual attention maps and online adaptable
  tracker,'' in \emph{2013 14th international workshop on image analysis for
  multimedia interactive services (WIAMIS)}.\hskip 1em plus 0.5em minus
  0.4em\relax IEEE, 2013, pp. 1--4.

\bibitem{sobral2015double_saliency4}
A.~Sobral, T.~Bouwmans, and E.-h. ZahZah, ``Double-constrained rpca based on
  saliency maps for foreground detection in automated maritime surveillance,''
  in \emph{2015 12th IEEE International Conference on Advanced Video and Signal
  Based Surveillance (AVSS)}.\hskip 1em plus 0.5em minus 0.4em\relax IEEE,
  2015, pp. 1--6.

\bibitem{cane2016saliency_saliency5}
T.~Cane and J.~Ferryman, ``Saliency-based detection for maritime object
  tracking,'' in \emph{Proceedings of the IEEE conference on computer vision
  and pattern recognition workshops}, 2016, pp. 18--25.

\bibitem{akilan2019sendec}
T.~Akilan and Q.~J. Wu, ``sendec: An improved image to image cnn for foreground
  localization,'' \emph{IEEE Transactions on Intelligent Transportation
  Systems}, vol.~21, no.~10, pp. 4435--4443, 2019.

\bibitem{wang2013stereovision}
H.~Wang and Z.~Wei, ``Stereovision based obstacle detection system for unmanned
  surface vehicle,'' in \emph{Proc. 2013 IEEE Int. Conf. on Robotics and
  Biomimetics}, 2013, pp. 917--921.

\bibitem{muhovic2019obstacle}
J.~Muhovi{\v{c}}, R.~Mandeljc, B.~Bovcon, M.~Kristan, and J.~Per{\v{s}},
  ``Obstacle tracking for unmanned surface vessels using 3-d point cloud,''
  \emph{IEEE Journal of Oceanic Engineering}, 2019.

\bibitem{lee2018image}
S.-J. Lee, M.-I. Roh, H.-W. Lee, J.-S. Ha, and I.-G. Woo, ``Image-based ship
  detection and classification for unmanned surface vehicle using real-time
  object detection neural networks,'' in \emph{The 28th International Ocean and
  Polar Engineering Conference}.\hskip 1em plus 0.5em minus 0.4em\relax
  International Society of Offshore and Polar Engineers, 2018.

\bibitem{moosbauer2019benchmark_smd_instance}
S.~Moosbauer, D.~Konig, J.~Jakel, and M.~Teutsch, ``A benchmark for deep
  learning based object detection in maritime environments,'' in
  \emph{Proceedings of the IEEE Conference on Computer Vision and Pattern
  Recognition Workshops}, 2019, pp. 0--0.

\bibitem{yang2019surface}
J.~Yang, Y.~Li, Q.~Zhang, and Y.~Ren, ``Surface vehicle detection and tracking
  with deep learning and appearance feature,'' in \emph{2019 5th International
  Conference on Control, Automation and Robotics (ICCAR)}.\hskip 1em plus 0.5em
  minus 0.4em\relax IEEE, 2019, pp. 276--280.

\bibitem{paszke2016enet}
A.~Paszke, A.~Chaurasia, S.~Kim, and E.~Culurciello, ``Enet: A deep neural
  network architecture for real-time semantic segmentation,'' \emph{arXiv
  preprint arXiv:1606.02147}, 2016.

\bibitem{badrinarayanan2017segnet}
V.~Badrinarayanan, A.~Kendall, and R.~Cipolla, ``Segnet: A deep convolutional
  encoder-decoder architecture for image segmentation,'' \emph{IEEE
  Transactions on Pattern Analysis and Machine Intelligence}, vol.~39, no.~12,
  pp. 2481--2495, 2017.

\bibitem{zhao2017pyramid}
H.~Zhao, J.~Shi, X.~Qi, X.~Wang, and J.~Jia, ``Pyramid scene parsing network,''
  in \emph{Proceedings of the IEEE conference on computer vision and pattern
  recognition}, 2017, pp. 2881--2890.

\bibitem{lin2017refinenet}
G.~Lin, A.~Milan, C.~Shen, and I.~Reid, ``Refinenet: Multi-path refinement
  networks for high-resolution semantic segmentation,'' in \emph{Proceedings of
  the IEEE Conference on Computer Vision and Pattern Recognition}, 2017, pp.
  1925--1934.

\bibitem{yu2018bisenet}
C.~Yu, J.~Wang, C.~Peng, C.~Gao, G.~Yu, and N.~Sang, ``Bisenet: Bilateral
  segmentation network for real-time semantic segmentation,'' in \emph{Proc.
  European Conf. Computer Vision}, 2018, pp. 325--341.

\bibitem{chen2018deeplab3}
L.-C. Chen, Y.~Zhu, G.~Papandreou, F.~Schroff, and H.~Adam, ``Encoder-decoder
  with atrous separable convolution for semantic image segmentation,'' in
  \emph{Proc. European Conf. Computer Vision}, 2018, pp. 801--818.

\bibitem{xu2020salmnet}
X.~Xu, T.~Yu, X.~Hu, W.~W. Ng, and P.-A. Heng, ``Salmnet: A structure-aware
  lane marking detection network,'' \emph{IEEE Transactions on Intelligent
  Transportation Systems}, 2020.

\bibitem{KristanCYB2015}
M.~Kristan, V.~S. Kenk, S.~Kovačič, and J.~Perš, ``Fast image-based obstacle
  detection from unmanned surface vehicles,'' \emph{IEEE transactions on
  cybernetics}, vol.~46, no.~3, pp. 641--654, 2016.

\bibitem{cane2018evaluating}
T.~Cane and J.~Ferryman, ``Evaluating deep semantic segmentation networks for
  object detection in maritime surveillance,'' in \emph{2018 15th IEEE
  International Conference on Advanced Video and Signal Based Surveillance
  (AVSS)}.\hskip 1em plus 0.5em minus 0.4em\relax IEEE, 2018, pp. 1--6.

\bibitem{kim2019vision}
H.~Kim, J.~Koo, D.~Kim, B.~Park, Y.~Jo, H.~Myung, and D.~Lee, ``Vision-based
  real-time obstacle segmentation algorithm for autonomous surface vehicle,''
  \emph{IEEE Access}, vol.~7, pp. 179\,420--179\,428, 2019.

\bibitem{steccanella2020intcatch}
L.~Steccanella, D.~Bloisi, A.~Castellini, and A.~Farinelli, ``Waterline and
  obstacle detection in images from low-cost autonomous boats for environmental
  monitoring,'' \emph{Robotics and Autonomous Systems}, vol. 124, p. 103346,
  2020.

\bibitem{bovcon2020wasr_icra}
B.~Bovcon and M.~Kristan, ``A water-obstacle separation and refinement network
  for unmanned surface vehicles,'' in \emph{Int. Conf. Robotics and
  Automation}.\hskip 1em plus 0.5em minus 0.4em\relax IEEE, 2020.

\bibitem{sun2020fuseseg}
Y.~Sun, W.~Zuo, P.~Yun, H.~Wang, and M.~Liu, ``Fuseseg: Semantic segmentation
  of urban scenes based on rgb and thermal data fusion,'' \emph{IEEE
  Transactions on Automation Science and Engineering}, 2020.

\bibitem{wang2019self}
H.~Wang, Y.~Sun, and M.~Liu, ``Self-supervised drivable area and road anomaly
  segmentation using rgb-d data for robotic wheelchairs,'' \emph{IEEE Robotics
  and Automation Letters}, vol.~4, no.~4, pp. 4386--4393, 2019.

\bibitem{luo2010pedestrian}
Y.~Luo, J.~Remillard, and D.~Hoetzer, ``Pedestrian detection in near-infrared
  night vision system,'' in \emph{2010 IEEE Intelligent Vehicles
  Symposium}.\hskip 1em plus 0.5em minus 0.4em\relax IEEE, 2010, pp. 51--58.

\bibitem{bovcon2018stereo}
B.~Bovcon, R.~Mandeljc, J.~Per{\v{s}}, and M.~Kristan, ``Stereo obstacle
  detection for unmanned surface vehicles by imu-assisted semantic
  segmentation,'' \emph{Robotics and Autonomous Systems}, vol. 104, pp. 1--13,
  2018.

\bibitem{geiger2012we_kitti}
A.~Geiger, P.~Lenz, and R.~Urtasun, ``Are we ready for autonomous driving? the
  kitti vision benchmark suite,'' in \emph{2012 IEEE Conference on Computer
  Vision and Pattern Recognition}.\hskip 1em plus 0.5em minus 0.4em\relax IEEE,
  2012, pp. 3354--3361.

\bibitem{cordts2016cityscapes}
M.~Cordts, M.~Omran, S.~Ramos, T.~Rehfeld, M.~Enzweiler, R.~Benenson,
  U.~Franke, S.~Roth, and B.~Schiele, ``The cityscapes dataset for semantic
  urban scene understanding,'' in \emph{Proceedings of the IEEE conference on
  computer vision and pattern recognition}, 2016, pp. 3213--3223.

\bibitem{Maddern2017}
W.~Maddern, G.~Pascoe, C.~Linegar, and P.~Newman, ``1 year, 1000 km: The oxford
  robotcar dataset,'' \emph{Int. Journal of Robotics Research}, 2017.

\bibitem{Barnes2020}
D.~Barnes, M.~Gadd, P.~Murcutt, P.~Newman, and I.~Posner, ``The oxford radar
  robotcar dataset: A radar extension to the oxford robotcar dataset,'' in
  \emph{ICRA}.\hskip 1em plus 0.5em minus 0.4em\relax IEEE, 2020.

\bibitem{VOT2019}
M.~Kristan, J.~Matas, A.~Leonardis, M.~Felsberg, R.~Pflugfelder, J.-K.
  Kamarainen, L.~Cehovin~Zajc, O.~Drbohlav, A.~Lukezic, A.~Berg, A.~Eldesokey,
  J.~Kapyla, G.~Fernandez, A.~Gonzalez-Garcia, A.~Memarmoghadam, A.~Lu, A.~He,
  A.~Varfolomieiev, A.~Chan, A.~Shekhar~Tripathi, A.~Smeulders,
  B.~Suraj~Pedasingu, B.~Xin~Chen, B.~Zhang, B.~Wu, B.~Li, B.~He, B.~Yan,
  B.~Bai, B.~Li, B.~Li, B.~Hak~Kim, and B.~Hak~Li, ``The seventh visual object
  tracking vot2019 challenge results,'' in \emph{Proceedings of the IEEE
  International Conference on Computer Vision Workshops}, 2019, pp. 0--0.

\bibitem{Dendorfer2020}
\BIBentryALTinterwordspacing
P.~Dendorfer, H.~Rezatofighi, A.~Milan, J.~Shi, D.~Cremers, I.~Reid, S.~Roth,
  K.~Schindler, and L.~Leal-Taixé, ``Mot20: A benchmark for multi object
  tracking in crowded scenes,'' in \emph{arXiv:2003.09003}, 2020. [Online].
  Available: \url{www.motchallenge.net}
\BIBentrySTDinterwordspacing

\bibitem{bovcon2019mastr}
B.~Bovcon, J.~Muhovi{\v{c}}, J.~Per{\v{s}}, and M.~Kristan, ``The mastr1325
  dataset for training deep usv obstacle detection models,'' in \emph{Int.
  Conf. Intell. Robots and Systems}.\hskip 1em plus 0.5em minus 0.4em\relax
  IEEE, 2019, pp. 3431--3438.

\bibitem{chan2020comprehensive}
Y.-T. Chan, ``Comprehensive comparative evaluation of background subtraction
  algorithms in open sea environments,'' \emph{Computer Vision and Image
  Understanding}, p. 103101, 2020.

\bibitem{ren2015faster}
S.~Ren, K.~He, R.~Girshick, and J.~Sun, ``Faster {R-CNN}: Towards real-time
  object detection with region proposal networks,'' in \emph{Advances in neural
  information processing systems}, 2015, pp. 91--99.

\bibitem{he2017mask}
K.~He, G.~Gkioxari, P.~Doll{\'a}r, and R.~Girshick, ``Mask r-cnn,'' in
  \emph{Proceedings of the IEEE international conference on computer vision},
  2017, pp. 2961--2969.

\bibitem{ma2019convolutional}
L.~Ma, W.~Xie, and H.~Huang, ``Convolutional neural network based obstacle
  detection for unmanned surface vehicle.'' \emph{Mathematical biosciences and
  engineering: MBE}, vol.~17, no.~1, pp. 845--861, 2019.

\bibitem{huang2017densely}
G.~Huang, Z.~Liu, L.~Van Der~Maaten, and K.~Q. Weinberger, ``Densely connected
  convolutional networks,'' in \emph{Proceedings of the IEEE Conference on
  Computer Vision and Pattern Recognition}, 2017, pp. 4700--4708.

\bibitem{bovcon2018iros}
B.~Bovcon and M.~Kristan, ``Obstacle detection for usvs by joint stereo-view
  semantic segmentation,'' in \emph{2018 IEEE/RSJ, International Conference on
  Intelligent Robots and Systems (IROS)}.\hskip 1em plus 0.5em minus
  0.4em\relax IEEE, 2018, pp. 5807--5812.

\bibitem{ronneberger2015u}
O.~Ronneberger, P.~Fischer, and T.~Brox, ``{U-Net}: Convolutional networks for
  biomedical image segmentation,'' in \emph{International Conference on Medical
  image computing and computer-assisted intervention}.\hskip 1em plus 0.5em
  minus 0.4em\relax Springer, 2015, pp. 234--241.

\bibitem{he2016resnet}
K.~He, X.~Zhang, S.~Ren, and J.~Sun, ``Deep residual learning for image
  recognition,'' in \emph{Proceedings of the IEEE conference on computer vision
  and pattern recognition}, 2016, pp. 770--778.

\bibitem{fefilatyev2006horizon_buoy_dataset}
S.~Fefilatyev, V.~Smarodzinava, L.~O. Hall, and D.~B. Goldgof, ``Horizon
  detection using machine learning techniques,'' in \emph{2006 5th
  International Conference on Machine Learning and Applications
  (ICMLA'06)}.\hskip 1em plus 0.5em minus 0.4em\relax IEEE, 2006, pp. 17--21.

\bibitem{mardct_dataset}
D.~D. Bloisi, L.~Iocchi, A.~Pennisi, and L.~Tombolini, ``{ARGOS-V}enice boat
  classification,'' in \emph{Advanced Video and Signal Based Surveillance
  (AVSS), 2015 12th IEEE International Conference on}, 2015, pp. 1--6.

\bibitem{seagull_dataset}
M.~Marques, P.~Dias, N.~Santos, V.~Lobo, R.~Batista, D.~Salgueiro, A.~Aguiar,
  M.~Costa, J.~da~Silva, A.~Ferreira, and J.~Sousa, ``Unmanned aircraft systems
  in maritime operations: Challenges addressed in the scope of the seagull
  project,'' in \emph{OCEANS 2015-Genova}.\hskip 1em plus 0.5em minus
  0.4em\relax IEEE, 2015, pp. 1--6.

\bibitem{ribeiro2017dataset_airborne_dataset}
R.~Ribeiro, G.~Cruz, J.~Matos, and A.~Bernardino, ``A dataset for airborne
  maritime surveillance environments,'' \emph{IEEE Transactions on Circuits and
  Systems for Video Technology}, 2017.

\bibitem{ipatch_dataset}
L.~{Patino}, T.~{Nawaz}, T.~{Cane}, and J.~{Ferryman}, ``Pets 2017: Dataset and
  challenge,'' in \emph{2017 IEEE Conference on Computer Vision and Pattern
  Recognition Workshops (CVPRW)}, July 2017, pp. 2126--2132.

\bibitem{smd_prasad2017video}
D.~K. Prasad, D.~Rajan, L.~Rachmawati, E.~Rajabally, and C.~Quek, ``Video
  processing from electro-optical sensors for object detection and tracking in
  a maritime environment: a survey,'' \emph{IEEE Transactions on Intelligent
  Transportation Systems}, vol.~18, no.~8, pp. 1993--2016, 2017.

\bibitem{marvel_dataset}
E.~Gundogdu, B.~Solmaz, V.~Y{\"u}cesoy, and A.~Ko{\c{c}}, ``Marvel: A
  large-scale image dataset for maritime vessels,'' in \emph{Asian Conference
  on Computer Vision}.\hskip 1em plus 0.5em minus 0.4em\relax Springer, 2016,
  pp. 165--180.

\bibitem{soloviev_detector_dataset}
V.~{Soloviev}, F.~{Farahnakian}, L.~{Zelioli}, B.~{Iancu}, J.~{Lilius}, and
  J.~{Heikkonen}, ``Comparing cnn-based object detectors on two novel maritime
  datasets,'' in \emph{2020 IEEE International Conference on Multimedia Expo
  Workshops (ICMEW)}, 2020, pp. 1--6.

\bibitem{dutta2016via}
A.~Dutta, A.~Gupta, and A.~Zissermann, ``{VGG} image annotator ({VIA}),''
  \url{http://www.robots.ox.ac.uk/~vgg/software/via/}, 2016.

\bibitem{dutta2019vgg}
\BIBentryALTinterwordspacing
A.~Dutta and A.~Zisserman, ``The {VIA} annotation software for images, audio
  and video,'' in \emph{Proceedings of the 27th ACM International Conference on
  Multimedia}, ser. MM '19.\hskip 1em plus 0.5em minus 0.4em\relax New York,
  NY, USA: ACM, 2019, pp. 2276--2279. [Online]. Available:
  \url{https://doi.org/10.1145/3343031.3350535}
\BIBentrySTDinterwordspacing

\bibitem{lin2014coco}
T.-Y. Lin, M.~Maire, S.~Belongie, J.~Hays, P.~Perona, D.~Ramanan,
  P.~Doll{\'a}r, and C.~L. Zitnick, ``Microsoft coco: Common objects in
  context,'' in \emph{European conference on computer vision}.\hskip 1em plus
  0.5em minus 0.4em\relax Springer, 2014, pp. 740--755.

\bibitem{gupta2019lvis}
A.~Gupta, P.~Dollar, and R.~Girshick, ``Lvis: A dataset for large vocabulary
  instance segmentation,'' in \emph{Proceedings of the IEEE Conference on
  Computer Vision and Pattern Recognition}, 2019, pp. 5356--5364.

\bibitem{long2015fully_segmentation_metrics}
J.~Long, E.~Shelhamer, and T.~Darrell, ``Fully convolutional networks for
  semantic segmentation,'' in \emph{Proceedings of the IEEE conference on
  computer vision and pattern recognition}, 2015, pp. 3431--3440.

\bibitem{dextr}
K.-K. Maninis, S.~Caelles, J.~Pont-Tuset, and L.~Van~Gool, ``Deep extreme cut:
  From extreme points to object segmentation,'' in \emph{Proceedings of the
  IEEE Conference on Computer Vision and Pattern Recognition (CVPR)}, June
  2018.

\bibitem{bochkovskiy2020yolov4}
A.~Bochkovskiy, C.-Y. Wang, and H.-Y.~M. Liao, ``Yolov4: Optimal speed and
  accuracy of object detection,'' \emph{arXiv preprint arXiv:2004.10934}, 2020.

\bibitem{tian2019fcos}
Z.~Tian, C.~Shen, H.~Chen, and T.~He, ``Fcos: Fully convolutional one-stage
  object detection,'' in \emph{Proceedings of the IEEE international conference
  on computer vision}, 2019, pp. 9627--9636.

\bibitem{liu2016ssd}
W.~Liu, D.~Anguelov, D.~Erhan, C.~Szegedy, S.~Reed, C.-Y. Fu, and A.~C. Berg,
  ``Ssd: Single shot multibox detector,'' in \emph{European conference on
  computer vision}.\hskip 1em plus 0.5em minus 0.4em\relax Springer, 2016, pp.
  21--37.

\bibitem{cheng2020panoptic}
B.~Cheng, M.~D. Collins, Y.~Zhu, T.~Liu, T.~S. Huang, H.~Adam, and L.-C. Chen,
  ``Panoptic-deeplab: A simple, strong, and fast baseline for bottom-up
  panoptic segmentation,'' in \emph{Proceedings of the IEEE/CVF conference on
  computer vision and pattern recognition}, 2020, pp. 12\,475--12\,485.

\bibitem{hazirbas2016fusenet}
C.~Hazirbas, L.~Ma, C.~Domokos, and D.~Cremers, ``Fusenet: Incorporating depth
  into semantic segmentation via fusion-based cnn architecture,'' in
  \emph{Asian conference on computer vision}.\hskip 1em plus 0.5em minus
  0.4em\relax Springer, 2016, pp. 213--228.

\bibitem{jegou2017one}
S.~J{\'e}gou, M.~Drozdzal, D.~Vazquez, A.~Romero, and Y.~Bengio, ``The one
  hundred layers tiramisu: Fully convolutional densenets for semantic
  segmentation,'' in \emph{Proceedings of the IEEE conference on computer
  vision and pattern recognition workshops}, 2017, pp. 11--19.

\bibitem{howard2017mobilenets}
A.~G. Howard, M.~Zhu, B.~Chen, D.~Kalenichenko, W.~Wang, T.~Weyand,
  M.~Andreetto, and H.~Adam, ``Mobilenets: Efficient convolutional neural
  networks for mobile vision applications,'' \emph{arXiv preprint
  arXiv:1704.04861}, 2017.

\bibitem{glorot2010understanding}
X.~Glorot and Y.~Bengio, ``Understanding the difficulty of training deep
  feedforward neural networks,'' in \emph{Proceedings of the thirteenth
  international conference on artificial intelligence and statistics}, 2010,
  pp. 249--256.

\bibitem{hinton2012neural}
G.~Hinton, N.~Srivastava, and K.~Swersky, ``Neural networks for machine
  learning,'' \emph{Coursera, video lectures}, vol. 264, p.~1, 2012.

\end{thebibliography}


\begin{IEEEbiography}
[{\includegraphics[width=1in,height=1in,keepaspectratio]{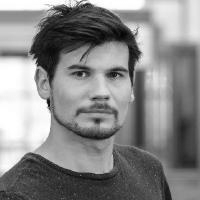}}]{Borja Bovcon}
received his M.Sc. degree from the Faculty of Mathematics and Physics at University of Ljubljana in 2017. He is currently working as a researcher at the ViCoS Laboratory, Faculty of Computer and Information Science, University of Ljubljana. His research interests are computer vision, obstacle detection and autonomous systems.
\end{IEEEbiography}

\begin{IEEEbiography}
[{\includegraphics[width=1in,height=1in,keepaspectratio]{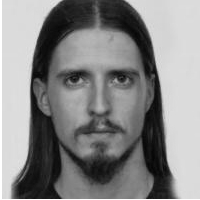}}]{Jon Muhovič}
received his M.Sc. degree from the Faculty of Computer Science and Informatics at University of Ljubljana in 2017. He is currently working as a researcher at the ViCoS Laboratory, Faculty of Computer and Information Science and at Laboratory for Machine Intelligence, Faculty of Electrical Engineering, University of Ljubljana. His research interests are computer vision, obstacle detection and autonomous systems.
\end{IEEEbiography}

\begin{IEEEbiography}
[{\includegraphics[width=1in,height=1in,keepaspectratio]{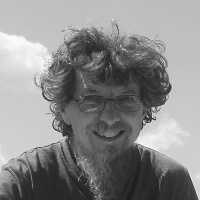}}]{Duško Vranac}
holds a diploma in physics from the University of Ljubljana and is one of the founders of the hydrographic survey company Harpha Sea, Koper. He was lead developer of the company's autonomous surface vehicles and has over 30 years experience in embedded systems, sea robotics, systems control, and navigation. Under his leadership the company developed a multi-sensor small unmanned surface vehicle for automated scanning and surveillance.
\end{IEEEbiography}

\begin{IEEEbiography}
[{\includegraphics[width=1in,height=1in,keepaspectratio]{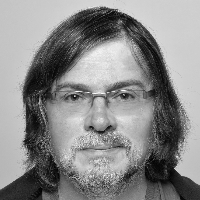}}]{Dean Mozetič}
holds a M.A. degree in astrophysics from Johns Hopkins University. He was a developer in the Harpha Sea d.o.o. department for autonomous surface vehicles, with over 20 years experience in embedded systems, sea robotics, systems control, and navigation. His recent work focuses on rapid bathymetric scanning for sea-bed model acquisition.
\end{IEEEbiography}

\begin{IEEEbiography}
[{\includegraphics[width=1in,height=1in,keepaspectratio]{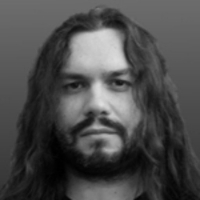}}]{Janez Perš}
received B.Sc., M.Sc., and Ph.D.\ degrees in Electrical Engineering at the Faculty of Electrical Engineering (FE), University of Ljubljana, in 1998, 2001, and 2004, respectively. He is currently an assistant professor at the Machine Vision Laboratory at the FE University of Ljubljana. His research interests lie in image-sequence processing, object tracking, human-motion analysis, dynamic-motion-based biometry, and in autonomous and distributed systems.
\end{IEEEbiography}

\begin{IEEEbiography}
[{\includegraphics[width=1in,height=1in,keepaspectratio]{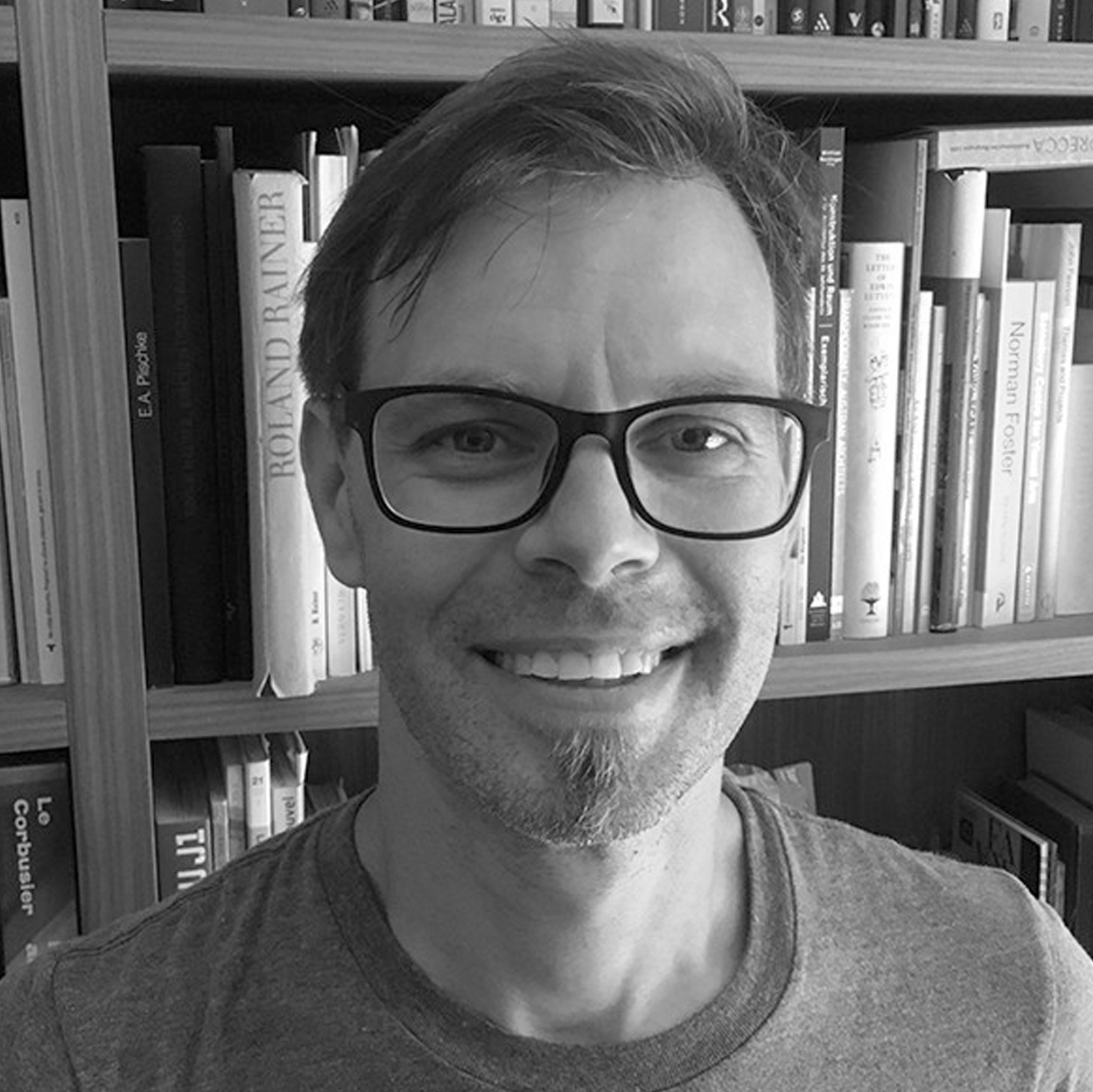}}]{Matej Kristan}
is currently an Associate Professor and a vice chair of the department of artificial intelligence at the Faculty of Computer and Information Science, University of Ljubljana. He leads the Visual object tracking VOT initiative, he is president of the IAPR Slovenian pattern recognition society and Associate Editor of IJCV. He has received fifteen research and teaching excellence awards, his works have been cited over 6400 times according to Google scholar and his h-index is 31. His research interests include visual object tracking, anomaly detection and segmentation, perception methods for autonomous boats and  machine-learning-based physics models. 
\end{IEEEbiography}

\end{document}